\documentclass{article}

\usepackage{arxiv}

\usepackage[utf8]{inputenc} 
\usepackage[T1]{fontenc}    
\usepackage{hyperref}       
\usepackage{nicefrac}       
\usepackage{microtype}      
\usepackage{lipsum}
\usepackage{graphicx}
\usepackage{amsmath,amssymb,amsfonts}%
\usepackage{mathrsfs}%
\usepackage[title]{appendix}%
\usepackage{xcolor}%
\usepackage{textcomp}%
\usepackage{manyfoot}%
\usepackage{algorithm}%
\usepackage{algorithmicx}%
\usepackage{algpseudocode}%
\usepackage{listings}%
\usepackage{subcaption}
\usepackage[numbers]{natbib}
\usepackage{tikz}
\usetikzlibrary{positioning, fit, shapes.geometric}
\usepackage[hypcap=false]{caption}
\usepackage[none]{hyphenat}
\usepackage{caption}
\usepackage{anyfontsize}
\DeclareCaptionFont{cap11}{\fontsize{12pt}{15pt}\selectfont}
\usepackage{array}     
\usepackage{booktabs,multirow,makecell,tabularx}
\usepackage{url}
\usepackage{placeins}   
\usepackage{xurl}        
\hypersetup{breaklinks=true}
\graphicspath{ {./images/} }

\title{Towards Efficient Real-Time Video Motion Transfer via Generative Time Series Modeling}

\author{
 Tasmiah Haque \\
  Department of Industrial and Management Systems Engineering\\
  West Virginia University\\
  Morgantown, WV 26505, USA \\
  \texttt{th00027@mix.wvu.edu} \\
   \And
 Md Asif Bin Syed \\
  Department of Industrial and Management Systems Engineering\\
  West Virginia University\\
  Morgantown, WV 26505, USA \\
  \texttt{ms00110@mix.wvu.edu} \\
  \And
 Byungheon Jeong \\
  Coupa Software\\
  San Mateo, CA, USA\\
  \texttt{joseph.jeong@coupa.com} \\
  \And
  Xue Bai \\
  Lyda Hill Department of Bioinformatics\\
  UT Southwestern Medical Center\\
  Dallas, TX 75235, USA\\
  \texttt{Xue.Bai@UTSouthwestern.edu}\\
  \And
  Sumit Mohan\\
  Intel Corporation\\
  Santa Clara, CA, USA\\
  \texttt{sumit.mohan@intel.com}\\
  \And
  Somdyuti Paul\\
  Department of Artificial Intelligence\\
  Indian Institute of Technology Kharagpur\\
  Kharagpur, West Bengal, India\\
  \texttt{somdyuti@cai.iitkgp.ac.in}
  \And
  Imtiaz Ahmed\\
  Department of Industrial and Management Systems Engineering\\
  West Virginia University\\
  Morgantown, WV 26505, USA \\
  \texttt{imtiaz.ahmed@mail.wvu.edu}\\
  \And
  Srinjoy Das \thanks{Corresponding author}\\
  School of Mathematical and Data Sciences\\
  Department of Industrial and Management Systems Engineering\\
  West Virginia University\\
  Morgantown, WV 26505, USA \\
  \texttt{srinjoy.das@mail.wvu.edu}\\ 
}

\begin{document}
\maketitle
\begin{abstract}
{Motion Transfer is an Artificial Intelligence (AI) technique that synthesizes videos by transferring motion dynamics from a driving video to a source image. In this work we propose a deep learning-based framework to enable real-time video motion transfer which is critical for enabling bandwidth-efficient applications such as video conferencing, remote health monitoring, virtual reality interaction, and vision-based anomaly detection. This is done using keypoints which serve as semantically meaningful, compact representations of motion across time, and are extracted from every video frame via a self-supervised detector. To enable bandwidth savings during video transmission we perform forecasting of keypoints using two generative time series models—Variational Recurrent Neural Networks (VRNN) and Gated Recurrent Units with Normalizing Flows (GRU-NF)—enabling both single and diverse future prediction modes. The predicted keypoints are transformed into realistic video frames using an optical flow-based module paired with a generator network, thereby facilitating accurate video forecasting and enabling efficient, low-frame-rate video transmission. Based on the application this allows the framework to either generate a deterministic future sequence or sample a diverse set of plausible futures.
Experimental results across three benchmark video datasets using state-of-the-art quality
and diversity metrics for video animation and reconstruction tasks demonstrate
that VRNN achieves the best point-forecast fidelity (lowest MAE) in the majority of evaluated settings in applications requiring stable and accurate multi-step forecasting (e.g., video conferencing, remote patient monitoring) and is particularly competitive in higher-uncertainty, multi-modal settings. This is achieved by utilizing the superior reconstruction property of the Variational Autoencoder and by introducing recurrently conditioned stochastic latent variables that carry past contexts to capture uncertainty and temporal variation. On the other hand the GRU-NF model enables richer diversity of generated videos while maintaining high visual quality to better support tasks like AI-driven anomaly detection. This is realized by learning an invertible,
exact-likelihood mapping between the keypoints and their latent representations which supports rich and controllable sampling of diverse
yet coherent keypoint sequences. Our work lays the foundation for next-generation AI systems that require real-time, bandwidth-efficient, and semantically controllable video generation, with broad implications for communication, health, and manufacturing applications.}
\end{abstract}



\section{Introduction}\label{sec1}

With the growing demand for ubiquitous services, telecommuting, immersive visual interactions, and realistic simulations, real-time video based applications such as video conferencing, augmented reality (AR), virtual reality (VR) gaming, and remote medical monitoring are becoming increasingly widespread \cite{Chan2019, Wang2021, Yang2022}. These applications demand not only high-quality visual outputs and seamless user interaction but also low-latency processing and efficient bandwidth usage—challenges that conventional video streaming and compression techniques struggle to meet effectively. Video motion transfer \cite{Wang2021}, an emerging Artificial Intelligence (AI) technique presents an effective and promising approach for realizing such applications by 
transferring motion from a driving video to a source frame or video, thereby enhancing user experiences and enabling more creative and engaging interactions. A key challenge in this space lies in modeling and predicting fine-grained  motion or object dynamics in a way that preserves both temporal coherence and visual fidelity, while also being bandwidth-efficient for deployment in real-time, edge computing environments. Pixel-based methods if used for prediction can suffer from high computational cost, error accumulation, and limited generalization in dynamic and diverse environments. Recent advances in representation learning and generative modeling have started to address these limitations by exploring latent spaces, structured motion representations, and probabilistic forecasting \cite{Oprea2020}. 
Among various representation learning based techniques for video synthesis and prediction, keypoint-based methods stand out by extracting keypoints from source and driving videos for enabling precise motion transfer \cite{ siarohin2019animating, siarohin2019first} while achieving superior video quality and compression beyond traditional codec-based methods. This not only reduces computational and bandwidth overhead, but also enables fine-grained control over motion dynamics, making keypoint-based methods ideal for real-time applications as compared to competing approaches which also claim to perform motion synthesis with high fidelity \cite{Pondaven2024}. 

A critical advantage of using keypoint-based motion transfer architectures is their ability to provide a natural, low-dimensional and interpretable structure for incorporating temporal prediction using generative time series models \cite{Chung2015, Rasul2020}, thereby enabling generalization across a variety of subjects and contexts. 
Unlike dense optical flow fields \cite{lu2023transflow}, which require significant storage and processing
per frame, keypoints offer a sparse yet semantically meaningful abstraction of
motion that reduces bandwidth and computational complexity. These are key
system requirements for our targeted real-time applications (video conferencing,
telehealth, Virtual Reality (VR) gaming, and manufacturing industry) where inference must run on resource-constrained edge platforms. Furthermore, while explicit pose parameters \cite{han2024prototypical} focus on specific human joints or body parts, keypoints detected in our pipeline are object-agnostic and self-supervised, allowing for broader applicability across human and non-human datasets. In contrast to pixel-level approaches, which suffer from error accumulation and sensitivity to occlusion or background clutter, keypoints focus on salient motion-driving features, enabling more stable and interpretable temporal modeling. By combining keypoint-based representation learning with deep generative forecasting models, real-time motion transfer frameworks can serve as the foundation for bandwidth-aware video AI systems, some examples of which are provided below.


\begin{itemize}
    \item \textbf{Video Conferencing:} In streaming applications such as video conferencing, keypoint-based motion transfer enables efficient future frame prediction by transmitting keypoints of initial frames from a source mobile or client, allowing the pipeline to predict and generate future frames at the receiving end \cite{Wang2021}. 
    Such an architecture not only reduces the dependency on ubiquitous network connectivity, but also optimizes computational efficiency and  bandwidth utilization.
\end{itemize}

\begin{itemize}
    \item \textbf{Virtual Reality (VR) Gaming:} For immersive experiences in VR gaming, it is crucial to transfer the physical actions of human players to virtual avatars. Anonymizing human players and generating seamless animations can be achieved by creating diverse sequences in a real-time motion transfer pipeline, thereby introducing natural variations in motion patterns of VR characters. In this manner it is possible to maintain realistic interactions in a  multiplayer environment, and players can feel more engaged without experiencing network related latencies. 
\end{itemize}

\begin{itemize}
    \item \textbf{Telehealth:} 
    For certain critical healthcare conditions it is essential for healthcare providers along with supporting personnel such as interns to monitor patients remotely. 
    However, in such cases patients may be hesitant to reveal their identity in virtual settings \cite{Yang2022} and this could also be mandated by regulatory restrictions. Keypoint-based motion transfer 
    enables future frame forecasting with anonymization that ensures both data privacy and accurate transmission of critical body movements of the patient for effective remote monitoring.
\end{itemize}

\begin{itemize}
    \item \textbf{Manufacturing Industry:} A key technical challenge when deploying Artificial Intelligence (AI)-driven inspection systems in manufacturing is the scarcity of comprehensive real-world datasets that include diverse defect types. Such limitations hinder the effectiveness of machine learning models, potentially resulting in missed detections or false negatives if predictions do not correspond precisely to known defects. In vision-based anomaly detection, by forecasting future video frames of units on the production line, the system can anticipate and visualize the onset of potential defects in upcoming frames. Generating multiple plausible future sequences instead of  a single prediction increases the chances of capturing visual patterns that may correspond to defects. This approach facilitates the generation of early and reliable defect warnings during real-time inspections. Moreover, the synthetic defect datasets generated through this approach expand the training data available for downstream classification or defect detection algorithms, enhancing their exposure to a wider variety of defect scenarios and improving overall robustness and detection accuracy.
\end{itemize}

In this work we go beyond pre-existing keypoint-based motion transfer frameworks such as the First Order Motion Model (FOMM) \cite{siarohin2019first} which provide spatial compression and focus on integrating probabilistic sequential generative models for keypoint forecasting in the motion transfer pipeline for realizing both spatial and temporal compression. In previous work \cite{Bai2024}, this has been demonstrated for applications involving video prediction using a Variational Recurrent Neural Network (VRNN) \cite{Chung2015} integrated within the FOMM pipeline. In our current work we substantially expand on these preliminary findings and also investigate other generative models including Normalizing Flow and Gated Recurrent Unit-Normalizing Flow (GRU-NF) \cite{Rasul2020} for keypoint forecasting within the motion transfer pipeline for both video prediction as well as diverse sample generation tasks. In the latter case we perform a systematic study of video diversity versus quality using a variety of metrics for measuring differences in generated realizations as well as their perceptual quality. 
Generative time series models are crucial here: unlike deterministic GRUs that typically assume simple output noise and collapse to a single trajectory, VRNNs introduce recurrently conditioned stochastic latents that carry history and uncertainty, while GRU-NF conditions an expressive flow on the recurrent state to model multi-modal futures without losing temporal coherence. By contrast, frame-wise VAEs/NFs without an autoregressive backbone ignore cross-time dependencies, whereas history-conditioned models such as VRNN and GRU-NF enable calibrated, diverse multi-step forecasts in multivariate settings.

Our enhanced architecture in this case can provide upto 20x bandwidth savings which is over the 10x savings that can be realized from existing keypoint-based motion transfer pipelines for video reconstruction and animation \cite{siarohin2019animating, siarohin2019first,Wang2021, Zhao2022}. To the best of our knowledge our work is the first systematic exploration of the potential of Normalizing Flow (NF) based architectures for predictive inference within a keypoint-based motion transfer framework. The rest of the paper is organized as follows. Section 2 discusses related works on video prediction, motion transfer and diverse sample generation. Section 3 describes our proposed real-time motion transfer architecture and the different 
forecasting models used in our work. The evaluation metrics used in this work to assess the performance of the models are explained in Section 4. Section 5 analyzes the results of our simulations based on the evaluation metrics for estimating the accuracy of generated videos and Section 6 provides a comparative analysis of the keypoint predictor models used. Finally, in Section 7 we present our conclusions and our plans for future work.

\section{Related Works}

In this section we discuss the evolution of video prediction techniques, highlighting recent advancements, particularly those that leverage deep learning models. We also review latent space sampling techniques from various generative models, 
evaluating their comparative ability to produce diverse, yet meaningful video samples. Additionally, various methods of video motion transfer and their feasibility in real-time applications are discussed.

\textbf{Single Sample  Video Prediction:} The prediction and synthesis of videos that accurately represent dynamic object movements is crucial for numerous practical applications. For this purpose various video prediction techniques including direct pixel synthesis, narrowing the prediction space to high-level feature spaces \cite{Oprea2020} have been explored by researchers. In order to accurately predict and synthesize videos by modeling object dynamics at the pixel level, numerous feature learning strategies have been explored, including adversarial training \cite{Luc2020} and gradient difference loss functions \cite{Mathieu2015}. 
Various recurrent and convolutional networks such as Convolutional Long-Short Term Memory (ConvLSTM) \cite{Shi2015}, E3d-LSTM \cite{Wang2018}, and  MSPred \cite{Villar-Corrales2022} are used for future frame prediction in raw pixel space. The common challenge faced by recurrent networks is vanishing gradient that hinders them to forecast in long term \cite{Gao2022}. To address this, transformer-based models incorporating self-attention mechanisms have been explored for long-term video prediction \cite{Shi2022, Ye2023, Lu2022}. While transformers have demonstrated strong performance on large text and image datasets, they are not yet well-optimized for fine-grained spatial-temporal modeling \cite{Ming2024}. To mitigate these challenges, a simpler video prediction model entirely based on convolutional neural networks (CNN), which outperformed both recurrent neural networks (RNNs) and Vision Transformers (ViTs) has been proposed in \cite{Gao2022}. The use of CNNs in video frame prediction has been explored by creating different types of models, i.e., Gradient
Difference Loss (GDL) \cite{Mathieu2015}, cascade convolution (PredCNN) \cite{Xu2018}, and spatially-displaced convolution (SDC-Net) \cite{Reda2018}. However, CNN-based frame synthesis is not suitable for spatially shifting pixels in input frames and is also inefficient for forecasting in large data sets \cite{Ming2024}. Generative Adversarial Network (GAN) models have also been used to predict a sequence of future frames based on a sequence of initial video frames using pixel space as input \cite{Aigner2018, Jang2018}. Additionally, flow-based architectures have been explored as an alternative that enables exact maximum likelihood learning utilizing invertible transformations. To perform forecasting, latent embeddings of video frames instead of raw pixels are used and future frames are reconstructed through an invertible mapping in \cite{Pottorff2019}. The key finding here is that NF based models outperform GANs that are based on adversarial training.


Inspite of their proliferation, pixel-based techniques face considerable challenges, especially for tasks requiring precise future predictions involving small or slow-moving objects. In addition, pixel-level prediction methods commonly accumulate significant errors due to inherent variability between consecutive frames. To address these challenges associated with high-dimensional pixel spaces, researchers have increasingly turned towards representation learning methods, such as semantic segmentation, human pose estimation, and keypoint-based representations. Before the widespread adoption of deep neural networks, traditional approaches employed Principal Component Analysis (PCA) or pose parameters for reconstructing video sequences with limited data, such as face boundaries \cite{Lopez1995, Koufakis1999}. Leveraging the combined strengths of VAEs and GANs, Walker et al. \cite{Walker2017} proposed an architecture that uses VAE to model possible future human poses in pixel space and then predicting future frames using a GAN with future poses used as conditional information. However, GANs often produce blurry and temporally inconsistent outputs \cite{Ming2024}. Similarly, VAEs  struggle to model complex video dynamics as a result of collapsing of posterior distribution to the prior \cite{Zhou2021}. Overall, pose-guided prediction methods, despite showing promise, have largely been restricted to videos containing human subjects \cite{Tang2019, Villegas2018, Villegas2017, Walker2017}. Semantic segmentation is another representation learning technique which instead of using raw pixel data extracts semantic elements by segmenting current frames and predict future optical flows through temporal models like Long Short-Term Memory (LSTM) networks \cite{Ranzato2014, Terwilliger2019, Hochreiter1997}. Keypoint-based approaches for representation learning, on the other hand, explicitly represent object-level dynamics, and can provide superior outcomes for trajectory prediction and action recognition. Recent research has demonstrated the efficacy of keypoints combined with reference frames for reconstructing future video frames, thereby substantially reducing accumulated errors inherent in pixel-space forecasting \cite{Minderer2019}.
In the context of using latent space embeddings, some recent works tackle the prediction challenge by using advanced modeling techniques instead of relying on basic linear motion predictions derived from past frames \cite{Ming2024}. In \cite{Esser2021}, a CNN-based vector quantization approach is introduced that discretizes images into a compact latent representation, enabling transformer-based autoregressive modeling for high-resolution image synthesis. Compressing high resolution videos into latent variables by using a vector quantized variational autoencoder (VQ-VAE) allowing for high resolution predictions has been proposed in \cite{Walker2021}. However, both the CNN and VQ-VAE based models often require extensive computational resources and substantial training data. Taking these limitations into account, our work employs keypoint-based prediction methodologies, integrating these representations effectively into video reconstruction and animation in a real-time motion transfer pipeline across multiple diverse datasets.

\textbf{Diverse Sample Video Prediction:} Incorporating uncertainty is another category of video prediction methods that allows generating diverse outcomes from a single input \cite{Oprea2020}. 
Video frames form multi-modal distributions in a high dimensional space and given this level of complexity, a range of approaches have incorporated latent variables into deterministic models e.g., LSTM and CNN to introduce stochasticity and variation in predictions. Gaussian Processes combined with LSTM have been used to enable the generation of diverse samples by encoding prior knowledge of future states from past observations, allowing for a more structured distribution over possible outcomes \cite{Shrivastava2021}. Latent variables have also been introduced into convolutional autoencoders to model video unpredictability through generative functions that capture diverse frame reconstructions \cite{Goroshin2015}. Similarly, convolutional recurrent networks have been trained to represent future stochasticity and generate varied sequences by sampling from learned latent distributions \cite{Hu2020}. Stochastic neural networks extend deterministic models by integrating Gaussian latent variables with convolutional encoders and spatial transformers, enabling the synthesis of multi-modal futures with enhanced diversity \cite{Fragkiadaki2017}. The stochastic variational video prediction framework advances this direction by extending the convolutional dynamic neural advection (CDNA) model to include latent variables, allowing for different possible futures per input sample \cite{Babaeizadeh2017}. Stochastic video generation model further improves uncertainty modeling by introducing a time-varying prior within an LSTM autoencoder, which better captures temporal dependencies in video sequences \cite{Denton2018}. Another common approach for generating diverse predictions is by sampling from the latent space of generative models such as VAE, GAN, and NF that can learn rich probabilistic representations of data \cite{Oprea2020}. To explore diverse video frame outcomes, GANs and VAEs are the most used generative models \cite{Xue2016, Yang2019, Sonderby2016}. The combination of GAN and VAE to leverage both stochastic and realistic generation is explored by \cite{Lee2018, Gur2020}. 
To utilize the direct optimization of data likelihood of normalizing flow architectures, \cite{Kumar2019} proposed VideoFlow model that outperforms state of the art VAE models in diverse possible video generation from a sequence of past observations. A trajectory forecasting model, diversity sampling for flow (DSF) that learns a joint distribution over multiple samples in the latent space of an NF has been developed in \cite{Ma2020} for improving both the quality and diversity of samples covering all possible modes of trajectories. 
While these latent variable-based methods enhance diversity, they often struggle to maintain high visual fidelity across generated frames \cite{Oprea2020}. In this work, we have used latent space sampling from VAE and NF based generative time series models, namely VRNN and GRU-NF for generating diverse video samples. 
We use metrics that jointly quantify diversity and reconstruction error w.r.t. the ground truth in order to evaluate the ability of different generative models to maintain visual fidelity while also producing diverse predictions.

\textbf{Video Motion Transfer:} Given the growing demand for high-quality interactive and immersive video experiences, particularly among users with limited bandwidth or connectivity, keypoint-based motion transfer models have emerged as an efficient solution for real-time object reconstruction and animation. These models enable a significant reduction in bandwidth requirements while maintaining video quality. Facial expressions and movements can be transferred between individuals using landmarks or keypoints combined with initial face embeddings. The Neural Talking Heads \cite{Zakharov2019} algorithm leverages meta-learning on extensive video datasets to generate talking head models from just a few or even a single image of a person using adversarial training. However, this approach struggles with landmark adaptation, leading to degraded performance when landmarks are extracted from different individuals. These shortcomings are addressed by introducing a keypoint-based model that synthesizes talking head videos by integrating a source image for appearance with a driving video for motion in \cite{Wang2021}. This method allows seamless video conferencing while minimizing bandwidth consumption without noticeable degradation in video quality. Monkey-Net proposed in \cite{siarohin2019animating} pioneered object-agnostic deep learning for motion transfer by enabling the animation of a source image using keypoints which are extracted from driving videos in a self-supervised manner. 
However, Monkey-Net relies on a zeroth-order keypoint model, leading to lower video quality when significant pose changes occur. To overcome this limitation, the First Order Motion Model (FOMM) \cite{siarohin2019first} was introduced, which decouples appearance and motion information using local affine transformations and self-learned keypoints. Expanding on these models, an end-to-end unsupervised framework, Thin Plate Spline model (TPS) developed in \cite{Zhao2022} is tailored to handle substantial pose differences between source and driving images. This approach incorporates thin-plate spline motion estimation to enhance optical flow adaptability and utilizes multi-resolution occlusion masks to reconstruct missing features more effectively, resulting in higher-quality image synthesis. 
The main limitations of FOMM arise under severe occlusions, lighting shifts, and highly non-rigid motion, where locally affine warps around sparse keypoints can underfit geometry. 
Although the TPS \cite{Zhao2022} addresses this, the 
added enhancements make the model heavier than FOMM’s 
locally affine pipeline, which can reduce headroom for real-time budgets on edge processors. LivePortrait \cite{guo2024liveportrait} enhances keypoint based animation with landmark-guided keypoints and large-scale training; 
however, it is portrait-specific (the pipeline works with faces only) and is not purely unsupervised like FOMM since it relies on landmarks. 
Continuous Piecewise Affine based (CPAB)  \cite{wang2024continuous} uses a diffeomorphic continuous piecewise-affine motion space and introduce foundation model \cite{bommasani2021opportunities} based training enhancements such SAM \cite{kirillov2023segment} guided keypoint semantics and DINO \cite{tumanyan2022splicing} based structure alignment, which improves robustness to complex, out-of-distribution motion. However during inference the model requires the solution of global CPAB parameters from keypoints, 
making motion estimation computationally costlier than FOMM 
and therefore harder to operate on real-time platforms.

In addition to keypoint-based approaches, alternative approaches to motion representation include dense optical flow fields \cite{lu2023transflow}, prototype-based motion learners \cite{lu2024promotion}, and part-based explicit pose modeling \cite{han2024prototypical}. These models provide rich motion encoding, especially in structured environments such as human motion. However, they often rely on supervision, require expensive pretraining, and are less adaptable to non-human motion or real-time applications. Moreover, recent works like \cite{liu2020video} have shown that over-complex motion features may not generalize well in real-time driving applications, reinforcing our choice of lightweight representations. 
Recent advancements in video motion transfer have also included proposals based on diffusion-based models. MotionShop \cite{Yesiltepe2024} introduces Mixture of Score Guidance (MSG), which decomposes motion into separate components and operates directly on pretrained diffusion models without additional training. Similarly, MotionFlow \cite{Meral2024} utilizes cross-attention maps to facilitate motion transfer during inference, enabling more flexible and generalizable motion synthesis. Furthermore, DiTFlow \cite{Pondaven2024} optimizes Attention Motion Flow (AMF) to transfer motion by utilizing Diffusion Transformers (DiTs) for higher motion consistency. Although these diffusion-based techniques achieve high-fidelity motion synthesis, their dependence on iterative denoising and complex optimization processes results in significant computational overhead, making them less viable for real-time applications.

Crucially, our targeted applications as described in Section \ref{sec1} require: (i) semantic and spatial consistency tracking the same object across frames while it moves and (ii) geometric consistency preserving stable, physically plausible motion cues. A self-supervised keypoint detector directly supports these requirements \cite{thewlis2017unsupervised, zhang2018unsupervised} by learning to place points at locations which are most informative for predicting how things move, yielding a compact, robust motion signal that is well-suited to forecasting and real-time deployment on resource-constrained platforms. Compared to the other approaches described above, keypoint-based methods like FOMM are more computationally efficient, making them ideal for real-time video applications where low-latency motion transfer is crucial. Our study builds upon keypoint-based motion transfer methods such as the FOMM 
and demonstrates high generalizability, thereby enabling efficient bandwidth optimization across various keypoint-based architectures such as \cite{siarohin2019animating, siarohin2019first, Wang2021, Zhao2022}. This makes our proposed architecture particularly well-suited for applications such as video reconstruction and animation while ensuring real-time processing capabilities.

\section{Methodology}

This section provides a comprehensive overview of our real-time motion transfer pipeline and explores different keypoint prediction methods evaluated in this research.

\subsection{The Proposed Pipeline}

\begin{figure*}[t]
    \centering
    \includegraphics[width=\textwidth]{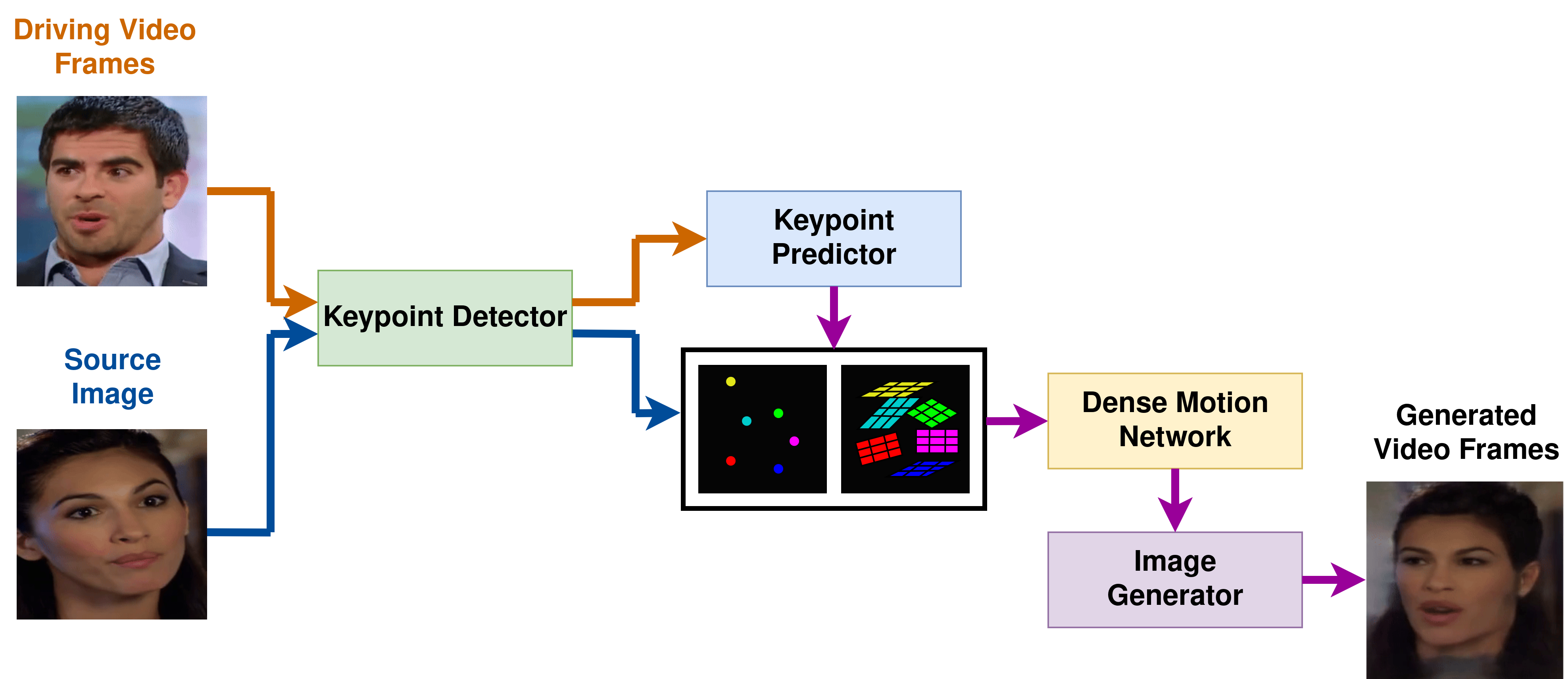}
    \caption{The proposed pipeline for real-time motion transfer (Single video sample prediction)}
    \label{fig:single pipeline}
\end{figure*}

\begin{figure*}[t]
    \centering
    \includegraphics[width=\textwidth]{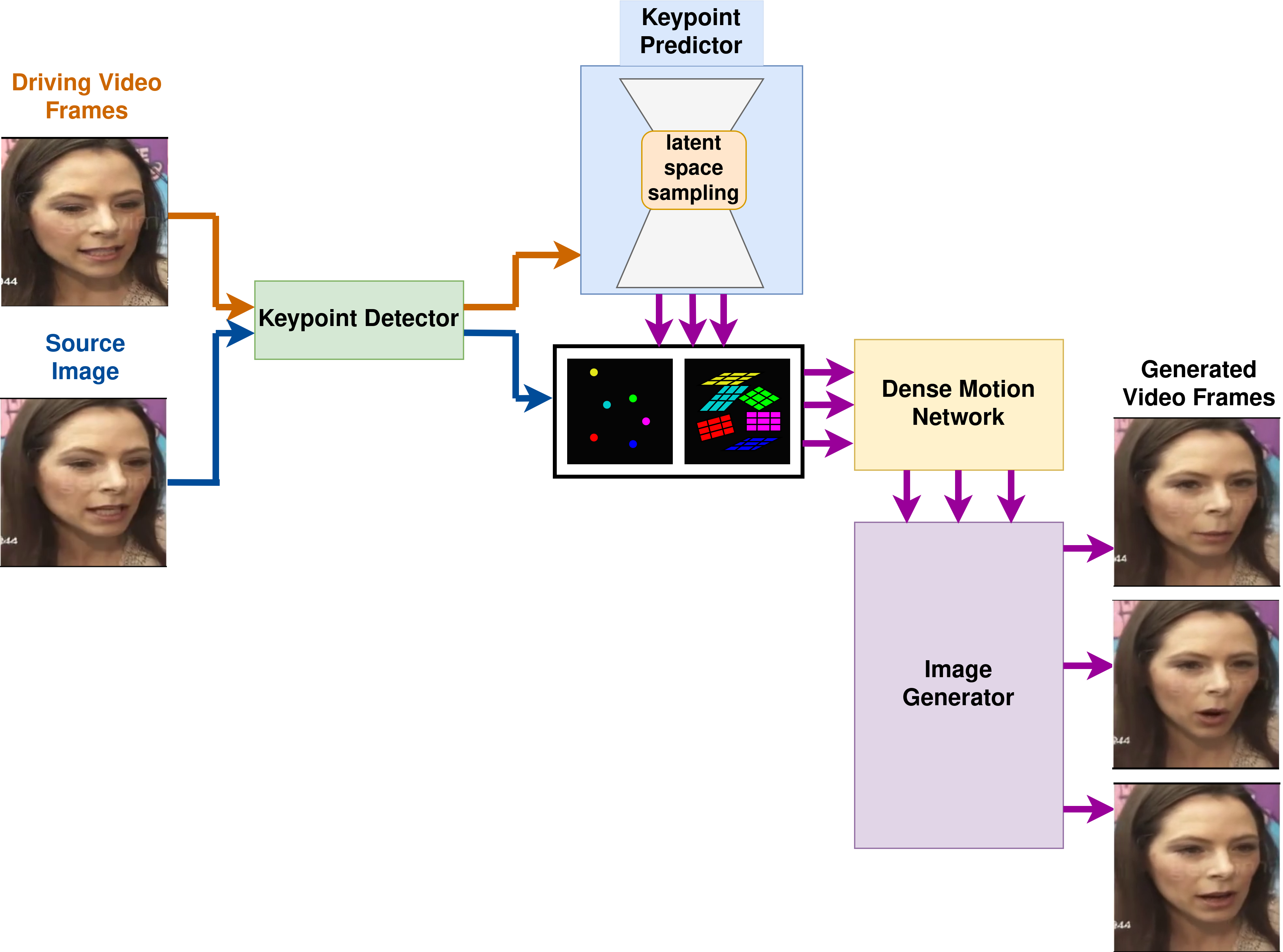}
    \caption{The proposed pipeline for real-time motion transfer (Diverse video sample prediction)}
    \label{fig:diverse pipeline}
\end{figure*}

Our real-time motion transfer proposal is based on the FOMM pipeline \cite{siarohin2019first} , although in general our method can be used for realizing bandwidth savings in other keypoint-based architectures such as \cite{Zhao2022, Wang2021}.
In the FOMM architecture, inputs to
the pipeline consist of a source image $\textbf{S} \in \mathbb{R}^{3 \times H \times W}$, where 3 denotes the number of channels corresponding to Red, Green, Blue (RGB), and $H$ and $W$ denote the height and width of the image respectively. and a sequence of $N$ driving video frames $\textbf{D} \in \mathbb{R}^{3 \times H \times W}$. At the start of the pipeline there is a keypoint detector which is a convolutional neural network that is used to produce $K$ feature maps from the source image {\textbf S} and each frame of the driving video {\textbf D}.
These extracted feature maps are then normalized and condensed to form keypoints by computing
the spatial expectation of the maps \cite{Minderer2019}. Each keypoint consists of the coordinates $x,y$ along with parameters of the local affine transformations around each keypoint. The latter set of features model the motion around a keypoint as a $2$x$2$ Jacobian matrix. In this work we set the hyperparameter $K=10$ which gives sufficiently good results. 
The 20 components representing the coordinates and the 40 components of the Jacobian matrix form a 60 dimensional time series.  
The keypoint predictor takes $N$ such keypoints corresponding to the driving video frames $\bf D$ and forecasts the next $M$ values. 
Following this the predicted keypoints go through the dense motion network which is used to align the keypoints captured from source image {\bf S} and driving video {\bf D}. The dense motion network is another convolutional neural network that generates heatmaps from the keypoints and then computes the dense motion field. The motion field is basically a function $\hat{\mathcal{T}}_{S \leftarrow D}$ which assigns every pixel position in {\bf D} to its equivalent position in {\bf S}. In addition, an occlusion mask $\hat{\mathcal{O}}_{S \leftarrow D}$ is generated by the dense motion network that is responsible for determining which image parts of {\bf D} must be reconstituted by warping from {\bf S} and which sections should be infilled based on the context. Finally, the generation module synthesizes the source image {\bf S} moving in accordance with the driving video {\bf D}. To achieve this, we utilize a generator network that warps the source image based on $\hat{\mathcal{T}}_{S \leftarrow D}$ and inpaints the occluded regions that are not visible in the source image \cite{siarohin2019first}. In this paper we demonstrate the use of the real-time motion transfer pipeline for multi-step ahead single video frame prediction as well as generating diverse samples of future video frames. The proposed pipeline for single and diverse sample prediction are demonstrated in Figures \ref{fig:single pipeline} 
and \ref{fig:diverse pipeline}.

Our implementation enables two types of video forecasting using real-time motion transfer: \textbf{reconstruction mode} and \textbf{transfer mode}. In reconstruction mode, the source image {\bf S} and each frame of the driving video {\bf D} come from the same video. Suppose a video $A$ consists of $N$ frames; ${f_1, f_2, ..., f_N}$, then the source image is the first frame $f_1$ and the frames for driving video are ${f_1, f_2, ..., f_N}$. On the other hand, in transfer mode, the source image {\bf S} and each frame of the driving video {\bf D} come from two different videos. In this case the source frame can be the first frame $g_1$ from a sequence of video frames $B$ denoted as ${g_1, g_2, ..., g_N}$ and the driving frames ${f_1, f_2, ..., f_N}$ are from a different video, video $A$. In both reconstruction and transfer modes forecasting is performed based on first $N$ keypoints to generate the next set of $M$ keypoints which are then synthesized into corresponding future video frames through the dense motion network and generator.
The training of the real-time motion transfer pipeline is done in two steps. First the FOMM is trained end-to-end in \textbf{reconstruction mode} except keypoint prediction. Following this keypoints are extracted from the driving video \textbf{D} using the trained keypoint detector. Then, the keypoint predictor is trained to predict next $M$ keypoints given initial $N$ values. Inference on the full pipeline including the keypoint predictor is performed in both \textbf{reconstruction mode} and \textbf{transfer mode}. 

\subsection{Keypoint Prediction Models}
\label{dl_models}

We have implemented five types of deep learning models in our FOMM pipeline for keypoint prediction including:
\begin{itemize}
    \item Autoregressive model: Gated Recurrent Unit (GRU)
    \item Generative models: Variational Autoencoder (VAE) and Normalizing Flow (NF)
    \item Generative time series models: Variational Recurrent Neural Network (VRNN) and Gated Recurrent Unit with Normalizing Flow (GRU-NF)
\end{itemize}

GRU was proposed in \cite{cho2014gru} and is a variant of the Recurrent Neural Network (RNN) which is designed to perform forecasting of high dimensional data. 
The problem of keypoint prediction 
can also be addressed using Deep Generative Models such as Variational Autoencoder (VAE) \cite{kingma2013auto} and Normalizing Flow (NF) \cite{Rezende2015}. Both of these are generative models that learn a probabilistic representation of high-dimensional data. 
Although VAEs and NFs are not inherently architected to perform forecasting, in this work we adapt them for keypoint prediction by reconstructing a future value from a given input.
Our primary focus in this work is to demonstrate the forecasting capabilities of two generative time series models namely VRNN and GRU-NF. The VRNN model proposed by \cite{Chung2015} combines the strengths of VAEs and RNNs to model sequential data whereas the GRU-NF proposed by \cite{Rasul2020} combines a GRU with an NF. In the following subsections we review the mathematical setup for each of these five models and also discuss how they are trained in our real-time motion transfer pipeline for keypoint forecasting.

\subsubsection{Gated Recurrent Unit (GRU)}

 GRU is a deep neural network which is designed to effectively capture temporal dependencies in sequential data, making it well-suited for tasks such as forecasting and classification.  
 In our work, GRU is used for keypoint forecasting in feedback or autoregressive mode, i.e., a new prediction from the current timestep is used as the input in the next timestep. This approach allows the GRU to retain state information between successive prediction steps, making it useful for capturing both short and long range temporal dependencies. 
 
 Let an input sequence of keypoints be denoted as \(\boldsymbol {x} = (\boldsymbol{x}_1,\boldsymbol{x}_2,...,\boldsymbol{x}_t)\). At each timestep \( t \), the GRU maintains a hidden state \( \boldsymbol{h}_t \) that is updated based on the current input \( \boldsymbol{x}_t \) and the previous hidden state \( \boldsymbol{h}_{t-1} \). This update mechanism of the hidden states in GRUs incorporates gating units---specifically, an update gate \(\boldsymbol{z}_t\) and a reset gate \(\boldsymbol{r}_t\)---which control the flow of information through the network \cite{cho2014gru}. First, the reset gate is computed as:
\begin{equation}
\boldsymbol{r}_t = \sigma\left(W_r \boldsymbol{x}_t + U_r \boldsymbol{h}_{t-1}\right)
\end{equation}
\noindent where \(\sigma(\cdot)\) is the sigmoid function, and \(W_r\), \(U_r\) are learnable weight matrices. Using the reset gate, a candidate hidden state \(\tilde{\boldsymbol{h}}_t\) is then calculated as below:
\begin{equation}
\tilde{\boldsymbol{h}}_t = \tanh\left(W \boldsymbol{x}_t + U \left(\boldsymbol{r}_t \odot \boldsymbol{h}_{t-1}\right)\right)
\end{equation}
\noindent where \(\tanh(\cdot)\) is the hyperbolic tangent function, \(W\) and \(U\) are corresponding learnable weight matrices, and \( \odot \) is the hadamard (element-wise) product. 

\noindent Next the update gate is computed as follows:
\begin{equation}
\boldsymbol{z}_t = \sigma\left(W_z \boldsymbol{x}_t + U_z \boldsymbol{h}_{t-1}\right)
\end{equation}

\noindent where \(W_z\), \(U_z\) are learnable weight matrices.
Finally, the hidden state is estimated using the update gate \(\boldsymbol{z}_t\) and the candidate hidden state \(\boldsymbol{\tilde{h}}_{t}\)
as follows:
\begin{equation}
\boldsymbol{h}_t = \left(1 - \boldsymbol{z}_t\right) \odot \boldsymbol{h}_{t-1} + \boldsymbol{z}_t \odot \tilde{\boldsymbol{h}}_t
\end{equation}
This gating mechanism allows the GRU to selectively retain or update information from previous input. During prediction, 
the GRU maps its hidden states to a probability distribution $p$ of outputs \( \boldsymbol{x}_{t+1}\) as below:
\begin{equation}
p\left(\boldsymbol{x}_{t+1} \mid \boldsymbol{x}_{\leq t}\right) = g_\tau\left(\boldsymbol{h}_t\right)
\end{equation}
where \(g_\tau\) is a nonlinear function with learnable parameters \(\tau\) that transforms the hidden state \( \boldsymbol{h}_{t}\) to the corresponding output \( \boldsymbol{x}_{t+1}\). In our work, the GRU is trained to predict \(\boldsymbol{x}_{t+1}\) from \(\boldsymbol{x}_t\) by minimizing the mean squared error (MSE) loss for $N$ training samples between the predicted output \(\boldsymbol{\hat{x}}\) and its corresponding ground truth \( \boldsymbol{x}\) at time $t+1$ as follows:

\begin{equation}
    \mathcal{L}_{\text{MSE}} = \frac{1}{N} \sum_{i=1}^{N} (\hat{\boldsymbol{x}}_{t+1}^{(i)} - \boldsymbol{x}_{t+1}^{(i)})^2
\end{equation}

During inference, the GRU takes as input a sequence of values from timesteps \( t_1, \dots, t_n \), and recursively generates the next \( k \) future values using its predictions from previous timesteps as inputs in an autoregressive manner.


\subsubsection{Variational Autoencoder (VAE)}

\begin{figure}[!ht]
	\centering
		\includegraphics[width=0.9\columnwidth]{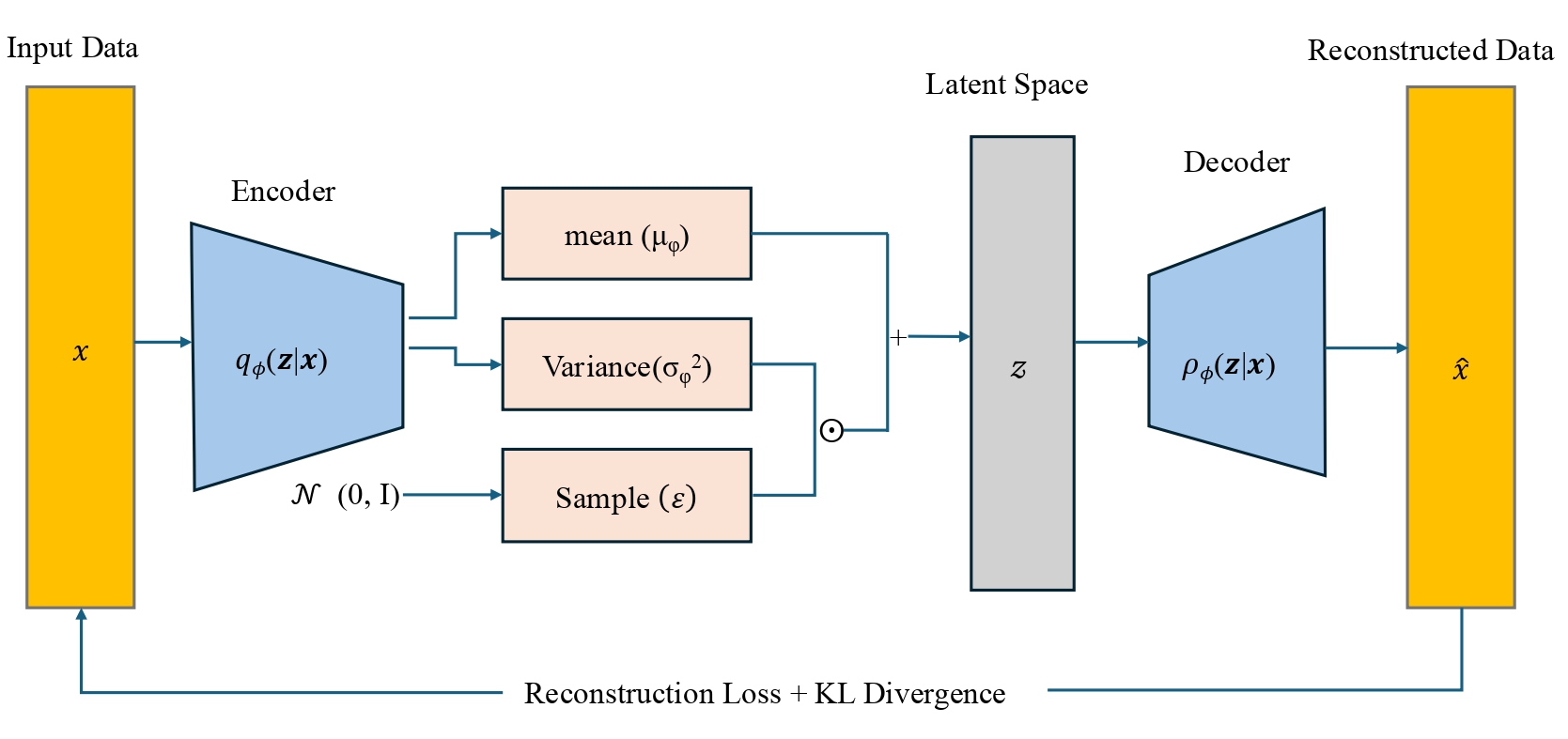}
    \caption{Variational Autoencoder (VAE) architecture with reparameterization trick.}
    \label{fig:vae_diagrams}
\end{figure}

The VAE is a deep generative model which is trained to minimize the reconstruction error between its inputs and outputs while simultaneously learning a structured, continuous latent space that enables smooth interpolation and sampling of new data. A VAE consists of three main components: an encoder that maps the input data to a distribution over latent variables, a latent space that captures uncertainty through probabilistic representations, and a decoder that reconstructs data using samples drawn from this latent distribution. For a time series of keypoints, a VAE encodes the input \( \boldsymbol{x}_t \) at each timestep \( t \) into a latent representation \( \boldsymbol{z}_t \) through a probabilistic encoder \( q_\phi \). The encoder estimates the posterior distribution of the latent variable \( \boldsymbol{z}_t \) as below:

\begin{equation}
q_\phi(\boldsymbol{z}_t | \boldsymbol{x}_t) = \mathcal{N}\left(\mu_\phi(\boldsymbol{x}_t), \sigma_\phi^2(\boldsymbol{x}_t)\right)
\end{equation}
where \( \mu_\phi(\boldsymbol{x}_t) \) and \( \sigma_\phi^2(\boldsymbol{x}_t) \) are the learned mean and variance of the Gaussian distribution. To enable backpropagation during training, the latent variable \( \boldsymbol{z}_t \) is sampled using the reparameterization trick \cite{rezende2014stochastic}:

\begin{equation}
\boldsymbol{z}_t = \mu_\phi(\boldsymbol{x}_t) + \sigma_\phi(\boldsymbol{x}_t) \odot \epsilon, \quad \epsilon \sim \mathcal{N}(0, I)
\end{equation}
where \( \epsilon \) is a noise vector drawn from a standard normal distribution and \( \odot \) denotes element-wise multiplication. The decoder \( p_\theta \), parameterized by weights \( \theta \), reconstructs the input \( \boldsymbol{x}_t \) from the latent variable \( \boldsymbol{z}_t \) by estimating the likelihood as below:

\begin{equation}
p_\theta(\boldsymbol{x}_t | \boldsymbol{z}_t) = \mathcal{N}\left(\mu_\theta(\boldsymbol{z}_t), \sigma_\theta^2(\boldsymbol{z}_t)\right)
\end{equation}
Here \( p_\theta(\boldsymbol{x}_t | \boldsymbol{z}_t) \) represents the likelihood of generating outputs ${\bf x}_t$ given the latent variable ${\bf z}_t$,  the decoder’s output is given by \( \mu_\theta(\boldsymbol{z}_t) \) and \( \sigma_\theta^2(\boldsymbol{z}_t) \) is the variance associated with the reconstruction. Figure \ref{fig:vae_diagrams} illustrates the architecture of a VAE. 

\noindent In our work, given $k > 1$ as the prediction horizon, the VAE is trained to predict \( \boldsymbol{x}_{t+k}\) from \( \boldsymbol{x}_{t}\) at time $t$ using the loss function as given below:

\begin{equation}
\mathcal{L}_{\text{VAE}} = \mathcal{L}_{\text{MSE}}  + KL[q_\phi(\boldsymbol{z}_t | \boldsymbol{x}_t) || p(\boldsymbol{z}_t)]
\end{equation}

\noindent where the MSE loss and KL-divergence loss functions are defined as follows:

\begin{equation}
\label{eqn: VAE recon}
    \mathcal{L}_{\text{MSE}} = \frac{1}{N} \sum_{i=1}^{N} (\hat{\boldsymbol{x}}_{t+k}^{(i)} - \boldsymbol{x}_{t+k}^{(i)})^2
\end{equation}

\begin{equation}
\begin{aligned}
KL(q_{\phi}(\bf{z}_t|\bf{x}_t)\,\|\,p(\bf{z}_t)) 
&= -\frac{1}{2N} \sum_{i=1}^{N} \sum_{j=1}^{J} \Big[ 
1 + \log\left(\sigma_{\phi,j}^{2}(\boldsymbol{x}_{t,i})\right)
- \mu_{\phi,j}^{2}(\boldsymbol{x}_{t,i}) - \sigma_{\phi,j}^{2}(\boldsymbol{x}_{t,i}) \Big]
\end{aligned}
\end{equation}

\noindent The first term in the loss function minimizes the reconstruction loss over $N$ training samples 
between the predicted output \(\boldsymbol{\hat{x}}\) and its corresponding ground truth \(\boldsymbol{x}\) at time $t + k$ while the second term, \( KL[\cdot || \cdot] \),  is the Kullback-Leibler divergence between the approximate posterior \( q_\phi(\boldsymbol{z}_t \mid \boldsymbol{x}_t) \) and the prior \( p(\boldsymbol{z}_t) \) 
where $\mathcal{N}(0, I)$ is typically chosen as the prior distribution over the latent space.

During inference, the VAE takes as input a sequence of values from timesteps \( t_1, \dots, t_n \), and generates the next \( k \) future values by reconstructing each \( \boldsymbol{x}_{k+i} \) from \(\boldsymbol{x}_i \), for \( i = 1, \dots, n \), where \(k>=n\), thereby shifting the input index forward with each prediction.

\subsubsection{Variational Recurrent Neural Network (VRNN)}
 
A VRNN is a generative time series model which combines the sequential modeling ability of RNNs with the probabilistic latent structure of VAEs. The VRNN model is essentially an RNN that contains a VAE at each time step \cite{Chung2015}. In order to maintain consistency among the various deep learning models in our work, we have used GRU as the RNN unit of the VRNN. The key steps describing how a VRNN works are descibed as follows.

For a time series of keypoints, the VRNN maintains a hidden state \( \boldsymbol{h}_t \) at each timestep \( t \), which is updated based on the input \( \boldsymbol{x}_t \), the 
prior \( \boldsymbol{z}_t \), and the previous hidden state \( \boldsymbol{h}_{t-1} \). The hidden state update in a VRNN is given as below:

\begin{equation}
\boldsymbol{h}_t = f\left(\varphi_\tau^{\boldsymbol{x}}(\boldsymbol{x}_t), \varphi_\tau^{\boldsymbol{z}}(\boldsymbol{z}_t), \boldsymbol{h}_{t-1}; \theta\right),
\end{equation}
where \( f \) represents a nonlinear activation function, and \( \varphi_\tau^{\boldsymbol{x}}(\cdot) \) and \( \varphi_\tau^{\boldsymbol{z}}(\cdot) \) are neural network functions parameterized by \( \tau \) for input and latent variable transformations respectively. In contrast with a VAE, the latent variable \( \boldsymbol{z}_t \) in case of a VRNN is sampled from a posterior distribution conditioned on both the input and the hidden state as below:

\begin{equation}
q(\boldsymbol{z}_t | \boldsymbol{x}_t, \boldsymbol{h}_{t-1}) = \mathcal{N}\left(\mu_{\boldsymbol{z}, t}, \sigma_{\boldsymbol{z}, t}^2\right),
\end{equation}
where \( [\mu_{\boldsymbol{z}, t}, \sigma_{\boldsymbol{z}, t}] = \varphi_\tau^{\text{enc}}(\varphi_\tau^{\boldsymbol{x}}(\boldsymbol{x}_t), \boldsymbol{h}_{t-1}) \). Here, \(\varphi_\tau^{\text{enc}}\) is the neural network for encoder, and \(\mu_{\boldsymbol{z},t}\) and \(\sigma_{\boldsymbol{z},t}\) are the parameters of the posterior distribution.

Similar to a VAE, the prior distribution in case of a VRNN over \( \boldsymbol{z}_t \) is also modeled as a Gaussian but in this case it depends on the previous hidden state of the GRU as below:

\begin{equation}
p(\boldsymbol{z}_t | \boldsymbol{h}_{t-1}) = \mathcal{N}\left(\mu_{0, t}, \sigma_{0, t}^2\right),
\end{equation}
where \( [\mu_{0, t}, \sigma_{0, t}] = \varphi_\tau^{\text{prior}}(\boldsymbol{h}_{t-1}) \). Here, \(\varphi_\tau^{\text{prior}}\) is the neural network for prior distribution, \(\mu_{\boldsymbol{0},t}\) and \(\sigma_{\boldsymbol{0},t}\) are the parameters of the prior distribution.
The generative process reconstructs \( \boldsymbol{x}_t \) from \( \boldsymbol{z}_t \) and \( \boldsymbol{h}_{t-1} \):

\begin{equation}
p(\boldsymbol{x}_t | \boldsymbol{z}_t, \boldsymbol{h}_{t-1}) = \mathcal{N}\left(\mu_{\boldsymbol{x}, t}, \sigma_{\boldsymbol{x}, t}^2\right),
\end{equation}
where \( [\mu_{\boldsymbol{x}, t}, \sigma_{\boldsymbol{x}, t}] = \varphi_\tau^{\text{dec}}(\varphi_\tau^{\boldsymbol{z}}(\boldsymbol{z}_t), \boldsymbol{h}_{t-1}) \). Here, \(\varphi_\tau^{\text{dec}}\) is the neural network for decoder, and \(\mu_{\boldsymbol{x},t}\) and \(\sigma_{\boldsymbol{x},t}\) are the parameters of the decoder output distribution.

\smallskip
\noindent The VRNN is trained by minimizing a loss function for a sequence of length \( T \) over $N$ training samples that 
comprises a reconstruction term and a regularization term (the KL divergence), which can be written as follows:

\begin{equation}
\begin{aligned}
\mathcal{L}_{\text{VRNN}} =
\frac{1}{N} \sum_{i=1}^N \Bigg[ \mathbb{E}_{q(\boldsymbol{z}_{\leq T}|\boldsymbol{x}_{\leq T})} \Bigg[ \sum_{t=1}^{T} \Big(-
KL\big(q(\boldsymbol{z}^{(i)}_t|\boldsymbol{x}^{(i)}_{\leq t}, \boldsymbol{z}^{(i)}_{<t}) 
\\ 
\|\, p(\boldsymbol{z}^{(i)}_t | \boldsymbol{x}^{(i)}_{<t}, \boldsymbol{z}^{(i)}_{<t})\big) 
+ \log p(\boldsymbol{x}^{(i)}_t \mid \boldsymbol{z}^{(i)}_{\leq t}, \boldsymbol{x}^{(i)}_{<t})
\Big) \Bigg]\Bigg]
\end{aligned}
\end{equation}

\noindent The first term measures the Kullback--Leibler divergence between the approximate posterior and the prior, thereby regularizing the latent space, while the second term represents the reconstruction loss, which encourages the decoder to generate keypoints similar to the input. 

\begin{figure}[!ht]
	\centering
		\includegraphics[width=0.6\columnwidth]{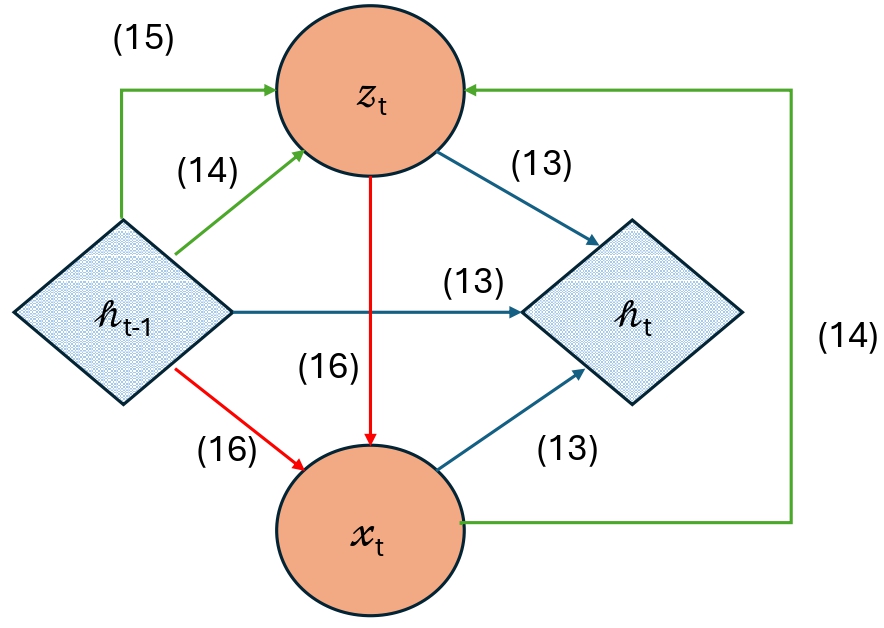}
    \caption{The graphical representation of VRNN describes the dependencies between the variables in Eqs.(13)-(16).The green arrows correspond to the computations involving the (conditional) prior and posterior on $z_t$. The Red arrows show the computations involving the generative network. The computations for $h_t$ are  shown with blue arrows.}
    \label{fig:vrnn_diagram}
\end{figure}

\noindent During inference, similar to a GRU, the VRNN takes as input a sequence of values from timesteps \( t_1, \dots, t_n \), and recursively generates the next \( k \) future values using its predictions from previous timesteps as inputs in an autoregressive manner. Figure \ref{fig:vrnn_diagram} provides a graphical representation of all the computations for a VRNN \cite{Ullah2020} . 



\subsubsection{Normalizing Flow (NF)}

Normalizing Flows (NF) are a class of generative models that transform a simple base distribution (e.g., Gaussian) into a complex target distribution using a sequence of differentiable and invertible mappings, enabling both efficient sampling and exact likelihood evaluation \cite{Papamakarios2021}. For the invertible mapping, NFs apply the `change of variables' rule \cite{Rezende2015} which is described as follows. Let \(\boldsymbol{x} \in \boldsymbol{X}\) be the observed data with a probability distribution \( p_{\boldsymbol{x}} \) which in our case is a time series of keypoints. Let the corresponding latent variable be \(\boldsymbol{z} \in \boldsymbol{Z}\) with a probability distribution \( p_{\boldsymbol{z}} \), and let g be an invertible and differentiable function where \(\boldsymbol{X} = g(\boldsymbol{Z})\) and f = \( g^{-1}\). Given the probability density function of \(\boldsymbol{z}\) , the change of variables formula for calculating the probability density function of \(\boldsymbol{x}\) is as follows:

\begin{equation}
p_X(\boldsymbol{x}) = p_Z(f(\boldsymbol{x})) \left| \det \left( \frac{\partial f(\boldsymbol{x})}{\partial \boldsymbol{x}^T} \right) \right|
\end{equation}

\begin{equation}
\log(p_X(\boldsymbol{x})) = \log \left( p_Z(f(\boldsymbol{x})) \right) + \log \left( \left| \det \left( \frac{\partial f(\boldsymbol{x})}{\partial \boldsymbol{x}^T} \right) \right| \right)
\end{equation}
where \(p_Z = \mathcal{N}(z; \mu = z, \sigma^2 = 1)
\) and \( \frac{\partial f(\boldsymbol{x})}{\partial \boldsymbol{x}^T} \) is the Jacobian of f at \(\boldsymbol{x}\).

For practical applications involving high dimensional data, estimating the Jacobian becomes computationally infeasible. To address these limitations, deep neural network (DNN) based estimation has been proposed for efficient and tractable computation of Jacobians. Real-valued Non-Volume Preserving (RealNVP) transformations, which use coupling layers as proposed by \cite{Dinh2016}, are a type of DNN-based approach that we use in our work to perform normalizing flow transformations. The equations of coupling layers used in RealNVP are described as follows.


Let \(\boldsymbol{x}\) be a D dimensional input data to the NF where $d < D$, and \(\boldsymbol{y}\) be the output of a coupling layer. The transformation performed by the coupling layer is defined by the following equations:

\begin{equation}
\label{eqn: nf split}
\boldsymbol{y}_{1:d} = \boldsymbol{x}_{1:d}
\end{equation}

\begin{equation}
\label{eqn: nf split 2}
\boldsymbol{y}_{d+1:D} = \boldsymbol{x}_{d+1:D} \odot \exp \left( s(\boldsymbol{x}_{1:d}) \right) + t(\boldsymbol{x}_{1:d})
\end{equation}

\noindent where \( \odot \) is the hadamard (element-wise) product, and s() and t() are scale and translation functions that are feedforward neural networks respectively from \( \mathbb{R}^d \rightarrow \mathbb{R}^{D - d} \). The Jacobian matrix becomes \(\exp \left( \sum_j s \left( \boldsymbol{x}_{1:d} \right)_j \right)\).

\noindent The inverse of the above functions are as follows:

\begin{equation}
\label{eqn: nf split inverse}
    \boldsymbol{x}_{1:d} = \boldsymbol{y}_{1:d}
\end{equation}

\begin{equation}
\label{eqn: nf split inverse 2}
    \boldsymbol{x}_{d+1:D} = \left( \boldsymbol{y}_{d+1:D} - t(\boldsymbol{y}_{1:d}) \right) \odot \exp \left( - s(\boldsymbol{y}_{1:d}) \right)
\end{equation}

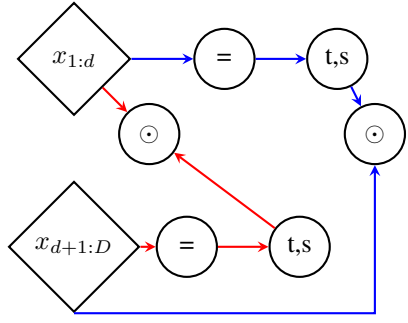
\begin{figure}
\begin{center}
\begin{tikzpicture}[node distance=1.5cm, thick, >=stealth]
    \tikzstyle{startstop} = [diamond, draw, fill=cyan!30, minimum size=1.5cm]
    \tikzstyle{process} = [circle, draw, fill=orange!30, minimum size=0.8cm]
    \tikzstyle{arrow} = [->]

    \node (n1) at (-2,2) [startstop] {$x_{1:d}$};
    \node (op1) at (-1,1) [process] {$\odot$};
    \node (eq1) at (0,2) [process] {=};
    \node (op2) [process, right of=eq1] {t,s};
    
    \node (n2) [startstop, below of=n1, yshift=-1cm] {$x_{d+1:D}$};
    \node (eq3) [process, right of=n2] {=};
    \node (op3) [process, right of=eq3] {t,s};
    \node (eq4) at (2,1) [process] {$\odot$};

    \draw [red,arrow] (n1) -- (op1);
    \draw [blue,arrow] (n1) -- (eq1);
    \draw [blue,arrow] (eq1) -- (op2);
    
    \draw [red,arrow] (n2) -- (eq3);
    \draw [red,arrow] (eq3) -- (op3);

    \draw [blue,arrow] (n2.south) -- ++(2,0) -- ++(2,0) -- (eq4.south);

    \draw [red,arrow] (op3) -- (op1);
    \draw [blue,arrow] (op2) -- (eq4);
    
\end{tikzpicture}
\end{center}
\caption{Alternating pattern of units in coupling layers of real-nvp}
\label{fig:NF_diagram}
\end{figure}

For splitting the input as shown in Eqs. \ref{eqn: nf split} and \ref{eqn: nf split 2}, channel-wise masking is applied in a shifting pattern \cite{Dinh2016}. A number of coupling layers are applied to map
\(\boldsymbol{x} \rightarrow \boldsymbol{y}_1 \rightarrow \cdots \rightarrow \boldsymbol{y}_{k-1} \rightarrow \boldsymbol{z}
\) all the while alternating the dimensions. The part that is identical in the current coupling layer is updated in the next one. Figure \ref{fig:NF_diagram} shows the shifting pattern of applying the coupling layer equations to the two parts of the input.

In our work, similar to the approach followed for the VAE,  given $k > 1$ as the prediction horizon, the NF is trained to predict \({\boldsymbol{x}_{t+k}}\) from \({\boldsymbol{x}_{t}}\) at time $t$ over $N$ training samples using the loss function as given below:

\begin{equation}
    \mathcal{L}_{NF} = \mathcal{L}_{\text{MSE}} + \mathcal{L}_{\text{NLL}}
\end{equation}

\noindent Here the MSE and negative log likelihood (NLL) loss functions are defined as below:





\begin{equation}
    \mathcal{L}_{\text{MSE}} = \frac{1}{N} \sum_{i=1}^{N} (\hat{\boldsymbol{x}}_{t+k}^{(i)} - \boldsymbol{x}_{t+k}^{(i)})^2
\end{equation}


\begin{equation}
    \mathcal{L}_{\text{NLL}} = - \frac{1}{N} \sum_{i=1}^N\log p_{\boldsymbol{x}}(\boldsymbol{x}^{(i)}_{t})
\end{equation}



\noindent The first term in the loss function minimizes the reconstruction loss over $N$ training samples 
between the predicted output \(\boldsymbol{\hat{x}}\) and its corresponding ground truth \(\boldsymbol{x}\) at time $t + k$ while the second term is the estimated NLL of the data.


During inference similar to a VAE, the NF takes as input a sequence of values from timesteps \( t_1, \dots, t_n \), and generates the next \( k \) future values by reconstructing each \({\boldsymbol{x}_{k+i}}\) from \({\boldsymbol{x}_{i}}\), for \( i = 1, \dots, n \), where \(k >= n\), thereby shifting the input index forward with each prediction.

\subsubsection{Gated Recurrent Unit-Normalizing Flow (GRU-NF)}

GRU-NF is a generative time series model proposed in \cite{Rasul2020} which combines the sequential modeling ability of GRUs with the exact likelihood modeling capability of NFs. The GRU-NF aims to effectively capture both the sequential structure and the distributional complexity of the underlying data. In our implementation, we use RealNVP as the NF component in the GRU-NF architecture. The key steps stating the details of how a GRU-NF works is described as follows.

For a time series of D dimensional keypoints ${\bf x}$, 
the GRU takes \(\boldsymbol{x}_{1:T}\) 
and generates a sequence of hidden states over this time window of $1:T$ as below:

\begin{equation}
\{\boldsymbol{h}_{1}, \ldots, \boldsymbol{h}_{T}\} = \text{GRU}(\{\boldsymbol{x}_{1}, \ldots, \boldsymbol{x}_{T}\}; \boldsymbol{h}_{0})
\end{equation}

\noindent Here the cold-start hidden state \(\boldsymbol{h}_{{0}}\) is set to  \(\vec{0}\). Note that since the hidden state \(\boldsymbol{h}\) contains the temporal dependencies 
from previous timesteps, it serves as a conditioning variable for the NF model to generate samples in its latent space as below.

\begin{equation}
\boldsymbol{z}_{1:T} =f(\boldsymbol{x}_{1:T}  | \boldsymbol{h}_{1:T})
\end{equation}



\noindent The following log-likelihood is maximized during training of the GRU-NF:

\begin{align}
\log p_{\boldsymbol{x}}(\boldsymbol{x}_{1:T} | \boldsymbol{h}_{1:T}; \theta) = \log p_{\boldsymbol{z}}(f(\boldsymbol{x}_{1:T} | \boldsymbol{h}_{1:T})) 
+ \log \left| \det J_f(\boldsymbol{x}_{1:T} | \boldsymbol{h}_{1:T}) \right|
\end{align}

\noindent where \(\theta\) denotes the entire set of parameters.

\noindent To calculate \(\log p_{\boldsymbol{x}}(\boldsymbol{x} | \mathbf{h}; \theta)\), the hidden state information from the GRU is concatenated to the inputs of the scaling and translation functions in the coupling layers of the NF i.e. \(
s(\mathrm{concat}(\boldsymbol{x}_{1:d}, \boldsymbol{h}))\) and \(t(\mathrm{concat}(\boldsymbol{x}_{1:d}, \boldsymbol{h}))\), where \( s(\cdot) \) and \( t(\cdot) \) represent the scaling and translation functions, respectively, and \(\boldsymbol{x}_{1:d} \) denotes part of the input vector where $d<D$.

\noindent The loss function of GRU-NF in terms of the number of training samples $N$ and the time window of size $T$ is as follows:

\begin{equation}
    \mathcal{L}_{GRU-NF} = -\frac{1}{NT} \sum_{i=1}^{N} \sum_{t=1}^T \log p_{\boldsymbol{x}}(\boldsymbol{x}_{t}^{(i)}| \boldsymbol{h}_{t}^{(i)}; \theta)
\end{equation}




\noindent During inference, the GRU is applied on the past input \(\boldsymbol{x}_{t-1}\) and hidden state \(\boldsymbol{h}_{t-1}\) from the previous time step as follows:

\begin{equation}
\boldsymbol{h}_{t} = \text{GRU}(\boldsymbol{y}_{t-1}; \boldsymbol{h}_{t - 1})
\end{equation}

\noindent Then, by sampling a noise vector \( \boldsymbol{z}_{t} \in \mathbb{R}^D \) from an isotropic Gaussian distribution, we apply the inverse flow to obtain a sample of the time series at the next time step:
\begin{equation}
{\boldsymbol{y}}_{t(NF)} = f^{-1}(\boldsymbol{z}_{t} \mid {\boldsymbol{h}}_{t})
\end{equation}

\begin{figure}[!ht]
\begin{center}
\begin{tikzpicture}
    \node[diamond, draw, fill=cyan!30, minimum size=1.2cm] (ht1) at (-2,2) {$h_{t-1}$};
    \node[circle, draw, fill=orange!30, minimum size=1.2cm] (xt1) at (-2,-1) {$y_{t-1}$};
    \node[diamond, draw, fill=cyan!30, minimum size=1.2cm] (ht) at (0.5,0.5) {$h_t$};
    \node[circle, draw, fill=orange!30, minimum size=1.2cm] (zt) at (5,2) {$z_t$};
    \node[circle, draw, fill=orange!30, minimum size=1.2cm] (xt) at (3,0.5) {$y_{t}$};
    \node[circle, draw, fill=orange!30, minimum size=1.2cm] (xtt) at (6,-1) {$y'_t$};

    \draw[red, ultra thick,->] (ht1) -- (ht) node[midway, above] {\textcolor{black}{}};
    \draw[red, ultra thick,->] (xt1) -- (ht) node[midway, left] {\textcolor{black}{}};
    \draw[green, ultra thick,->] (ht) -- (xt) node[midway, above] {\textcolor{black}{}};
    \draw[green, ultra thick,->] (zt) -- (xt) node[midway, right] {\textcolor{black}{}};
    \draw[blue, ultra thick,->] (xt) -- (xtt) node[midway, right] {\textcolor{black}{}};

    
\end{tikzpicture}
\end{center}
\caption{The flow diagram of GRU-NF architecture displays the dependencies between the variables in Eqs.(31)-(32). The red arrows show the GRU hidden state update, and the green arrows show the conditional sample generation with NF inverse transformation during inference.}
\label{fig:GRU-Nf_diagram}
\end{figure}
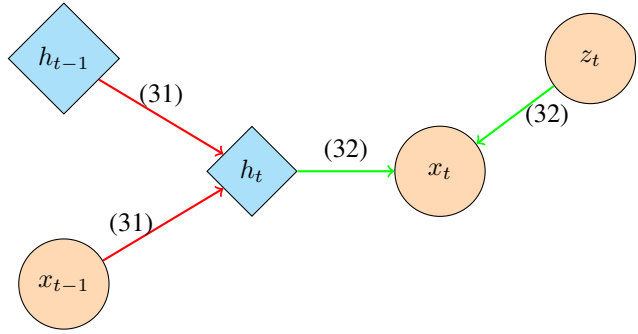

\noindent We then use this newly predicted output to compute the next hidden state \({\boldsymbol{h}}_{t+1} \) via the GRU, and repeat this process over the desired inference horizon. The flow diagram of the components of GRU-NF architecture is displayed in Figures \ref{fig:GRU-Nf_diagram_train} and \ref{fig:GRU-Nf_diagram}.




\begin{figure}[!ht]
\begin{center}
\begin{tikzpicture}
    \node[diamond, draw, fill=cyan!30, minimum size=1.2cm] (ht1) at (-2,2) {$h_{0}$};
    \node[circle, draw, fill=orange!30, minimum size=1.2cm] (xt1) at (-2,-1) {$x_{1:T}$};
    \node[diamond, draw, fill=cyan!30, minimum size=1.2cm] (ht) at (1,2) {$h_{1:T}$};
    \node[circle, draw, fill=orange!30, minimum size=1.2cm] (zt) at (3,-1) {$z_{1:T}$};

    \draw[red, thick,->] (ht1) -- (ht) node[midway, above] {\textcolor{black}{}};
    \draw[red, thick,->] (xt1) -- (ht) node[midway, left] {\textcolor{black}{}};
    \draw[green, thick,->] (ht) -- (zt) node[midway, above] {\textcolor{black}{}};
    \draw[green, thick,->] (xt1) -- (zt) node[midway, above] {\textcolor{black}{}};

    
\end{tikzpicture}
\end{center}
\caption{The flow diagram of GRU-NF architecture displays the dependencies between the variables in Eqs.(27)-(28). The red arrows show the GRU hidden state update, and the green arrows show the conditional sample generation with NF forward transformation during training.}
\label{fig:GRU-Nf_diagram_train}
\end{figure}
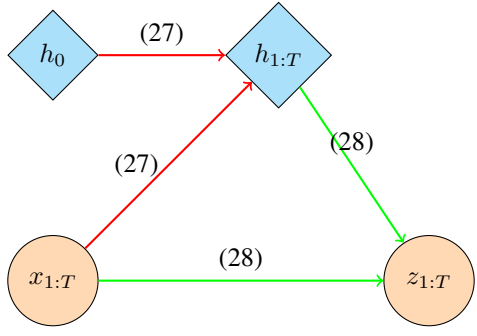

\section{Evaluation Metrics}

To compare accuracy of the predicted videos compared to ground truth as well as quantify the video diversity and quality, we have used the following metrics in this work:

\begin{itemize}
\item Mean Absolute Error (MAE) to assess spatial fidelity between predicted and real video sequences
\item JEPA (Joint Embedding Predictive Architecture) Embedding Distance (JEDi) to capture spatiotemporal consistency between predicted and real video sequences
\item Average Pair-wise Displacement (APD) to measure the diversity among the generated samples
\item Structural Similarity (SSIM) to evaluate the visual quality of generated samples against ground truth from human perspective
\end{itemize}
These metrics are described as below.

\subsection{Mean Absolute Error (MAE)}

Mean Absolute Error (MAE) is a widely used metric for assessing the accuracy of predictions in various applications, including video prediction \cite{siarohin2019first}. MAE quantifies the average magnitude of errors between predicted and actual pixel positions of video frames, without considering their direction. MAE is calculated as:
\begin{equation}
\text{MAE} = \frac{1}{P \cdot T} \sum_{p=1}^{P} \sum_{t=1}^{T} \|\hat{\mathbf{v}}_t^{(p)} - \mathbf{v}_t^{(p)}\|_1
\end{equation}

\noindent where \( P \) is the number of pixel values in a video frame, \( T \) is the number of timesteps, \( \hat{\mathbf{v}}_t^{(p)} \) is the predicted pixel value at time \( t \) and pixel \( p \) and \( \mathbf{v}_t^{(p)} \) is the corresponding ground truth position.

\subsection{JEPA Embedding Distance (JEDi)}

The JEPA (Joint Embedding Predictive Architecture) Embedding Distance (JEDi) \cite{luo2024beyond} offers a powerful approach for evaluating video generation quality by measuring the distributional similarity between generated and real video sequences. Addressing limitations of metrics like Fr\'echet Video Distance (FVD) \cite{unterthiner2018fvd}, which assumes Gaussianity of the underlying video data, JEDi does not require such distributional assumptions and  leverages Maximum Mean Discrepancy (MMD) with a polynomial kernel to compute the distance between JEPA-derived video feature sets. MMD is an integral probability metric which quantifies the difference between two probability distributions by finding a function within a chosen class that maximizes the difference in expectations over the distributions \cite{gretton2012kernel}. In addition to its applicability over a broad range of distributions, JEDi also aligns more closely with human evaluations compared to FVD and offers a scalable alternative to manually collecting human opinion scores, making it well-suited for large-scale evaluation of generative video models. The setup for JEDi estimation is described briefly as below.

Formally, with a function class $\mathcal{F}$, the MMD is defined as:
\begin{equation}
\mathrm{MMD}[\mathcal{F}, p, q] = \sup_{f \in \mathcal{F}} (\mathbb{E}_x[f(x)] - \mathbb{E}_y[f(y)])
\end{equation}
where $x \sim p$ and $y \sim q$, \(\sup_{f \in \mathcal{F}}\) denotes the maximum over all functions in the class $\mathcal{F}$, and $\mathbb{E}_x[f(x)]$ and $\mathbb{E}_x[f(y)]$ represent the expected values of $f(x)$ and $f(y)$ when $x$ and $y$ are drawn from distributions $p$ and $q$ respectively. In JEDi, the function class is implicitly defined by the polynomial kernel as follows:
\begin{equation}
    k(x,y) = (\langle x,y \rangle + c)^d
\end{equation}
where, k is the polynomial kernel, c is the bias term and d is the degree of the kernel. The kernel captures higher-order moments, making it particularly suitable for assessing temporal coherence in videos. 

Let, $Q$ be the distribution over JEPA-derived features from real videos and $R$ be the distribution over JEPA-derived features from generated videos.  JEDi is thus defined as:

\begin{equation}
\mathrm{JEDi}(Q,R) = \mathrm{MMD}_{poly}(X, Y)
\end{equation}

\noindent where $X = \{x_1,x_2,...,x_m\}$ and $Y = \{y_1,y_2,...,y_n\}$ represent sets of JEPA derived feature samples, with $m$ and $n$ samples drawn independently and identically distributed (i.i.d.) from distributions $Q$ and $R$, respectively. A lower JEDi score indicates a better alignment between generated and real video distributions.

\subsection{Average Pair-wise Displacement (APD)}
Average Pair-wise Displacement (APD) \cite{chen2024mixed} is designed to evaluate the diversity of predicted trajectories 
in multi-modal forecasting scenarios. 
In this work we use APD to measure diversity across videos generated using our real-time motion transfer pipeline. Unlike traditional metrics that compare each prediction to the ground truth, APD measures diversity by computing the average pairwise displacement across all predicted trajectories for a given video. In the context of diverse video prediction, it quantifies how much the generated video samples differ from one another over a time interval. APD is defined as:
\begin{equation}
 \mathrm{APD} = \frac{\sum_{i=1}^{M} \sum_{j=1}^{M} \left\|{}^{(i)}\hat{\mathbf{v}}_{t_{1}:t_{2}}^{a} - {}^{(j)}\hat{\mathbf{v}}_{t_{1}:t_{2}}^{a}\right\|_1}{M^2 \cdot (t_2 - t_1) \cdot P} 
\end{equation}

\noindent where $M$ is the number of generated samples for a given video, P is the number of pixel values in a video frame, and ${}^{(i)}\hat{\mathbf{v}}_{t_{1}:t_{2}}^{a}$ and ${}^{(j)}\hat{\mathbf{v}}_{t_{1}:t_{2}}^{a}$ denote the $i$-th and $j$-th predicted trajectory for a video $a$ over the time interval from $t_1$ to $t_2$, respectively. The numerator sums the Manhattan distances between every pair of predicted trajectories, while the denominator normalizes this total by the number of prediction pairs, the length of the prediction horizon and the number of pixel values in a frame. A higher APD value indicates that the predictions are more widely dispersed, reflecting a model’s ability to generate diverse future trajectories rather than merely converging on the most likely outcome. An APD value which is close to $0$ indicates that there is almost no diversity among the generated samples.

\subsection{Structural Similarity (SSIM)}

The Structural Similarity Index (SSIM) \cite{Wang2004SSIM} is a quality metric which is used for evaluating the perceptual quality of generated images. It quantifies the similarity between a generated frame and the corresponding ground truth frame by jointly comparing luminance, contrast, and structural information. It has been proven that for estimating video reconstruction quality from the perspective of the human visual system, SSIM is a more effective metric than the commonly used Peak Signal-to-Noise Ratio (PSNR) \cite{Kotevski2009}. The SSIM metrics lies between $0$ and $1$ where a value of $0$ indicates no similarity between generated video frame and ground truth video frame, and $1$ indicates 100\% similarity between the two. 

SSIM is defined as:
\begin{equation}
\text{SSIM}(g,r) = \frac{(2\mu_g\mu_r + c_1)(2\sigma_{gr} + c_2)}{(\mu_g^2 + \mu_r^2 + c_1)(\sigma_g^2 + \sigma_r^2 + c_2)}
\end{equation}
where g is the generated video frame and r is the real video frame, \(\mu_g\) and \(\mu_r\) represent the mean pixel intensities of frames \(g\) and \(r\), \(\sigma_g^2\) and \(\sigma_r^2\) denote their variances, and \(\sigma_{gr}\) is the covariance between the two frames. The constants \(c_1\) and \(c_2\) are included to stabilize the division in cases of denominator values close to $0$.

\section{Results}

We apply the five models discussed in Section \ref{dl_models} to three video datasets for evaluating single sample prediction in both reconstruction and transfer modes using our real-time motion transfer pipeline. For computational complexity reasons we evaluate diverse sample prediction using two datasets and two models namely the VRNN and GRU-NF in both reconstruction and transfer modes. 
Beyond computational considerations, VRNN and GRU-NF are chosen for their architectures, which are specifically designed to capture temporal dependencies in high-dimensional data such as keypoint time series. Moreover, both models leverage their latent spaces to generate multiple diverse samples at each timestep from a single input.

In this section, we introduce the datasets used in our evaluation following which the experimental results obtained by applying the models for single sample prediction and diverse sample prediction are presented and analyzed.

\subsection{Datasets}

The three datasets used in this paper are described below:

\textbf{MGIF dataset:} We employ 284 videos for training and 34 for testing from the MGIF dataset with 256 x 256 resolution \cite{siarohin2019animating}, which records periodic animal movements. These videos exhibit a range of frame counts, from 168 to 10,500 frames 
and allow for a thorough assessment of the model's capability to recognize distinct movement patterns among various animal species. 

\textbf{BAIR dataset:} For the BAIR dataset \cite{ebert2017self}, we work with 5001 training videos and 256 testing videos with 256 x 256 resolution, each consisting of  30 frames. These depict robotic arms interacting with different objects, offering a rigorous test of the model's effectiveness in understanding dynamic environments with complex backgrounds and irregular motions. 

\textbf{VoxCeleb dataset:} We utilize 3884 training videos and 44 testing videos from the VoxCeleb dataset  with 256 x 256 resolution \cite{nagrani2017voxceleb}, which includes interview clips featuring different celebrities. The video lengths vary with frame counts ranging from 72 to 1228 frames. This dataset is particularly valuable for assessing the model's proficiency in picking up on subtle facial expressions which is a critical requirement for applications like video conferencing where conveying nuanced emotions is important.

The MGIF and BAIR datasets serve as our non-human benchmarks. MGIF captures motion of deformable objects and animated characters, while BAIR consists of robotic arm manipulation sequences. Including these datasets allows us to evaluate the proposed models under non-human and highly deformable motion regimes, complementing the human-centric VoxCeleb dataset. 

\subsection{Single Video Sample Prediction}

We have applied the GRU, VAE, VRNN, NF and GRU-NF models to the three datasets in both reconstruction and transfer modes for single sample prediction. The prediction is performed using three different prediction horizons in order to observe how the models perform over varying horizon lengths. For each dataset the prediction horizons are generated using nonoverlapping sliding windows over the dataset. For both the MGIF and VoxCeleb datasets, we evaluate three prediction settings: 6–6, 6–24, and 12–12. In 6–6 and 6-24, the first 6 frames are used as input and the subsequent 6 frames or 24 frames are predicted respectively. In 12–12, the first 12 frames are used as input and the subsequent 12 frames are predicted. For the BAIR dataset, we evaluate 5–5, 5–20, and 15–15. In 5–5 and 5-20, the first 5 frames are used as input and the next 5 frames or 20 frames are predicted respectively. In 15–15, the first 15 frames are used as input and the next 15 frames are predicted.

For assessing single sample prediction in reconstruction and transfer modes, we have used the MAE and JEDi metrics. As there is no original video for transfer mode, ground truth videos are generated using the FOMM pipeline with the original keypoints of the driving video and source image. The results for both reconstruction and transfer mode for MGIF, BAIR and VoxCeleb dataset are demonstrated in Tables \ref{table:1}, \ref{table:2}, \ref{table:3}, \ref{table:4}, \ref{table:5} and \ref{table:6}. The first column of the tables represents the number of input frames and the number of output frames used for prediction. 

Overall, the results show a systematic dependence on dataset characteristics and prediction horizons, rather than a single model dominating all settings. In terms of pixel-level fidelity, VRNN achieves the best performance in the majority of cases (13 out of 18 settings), indicating strong reconstruction accuracy across datasets. Conversely, the JEDi metric which reflects spatiotemporal coherence tends to favor GRU. This implies that deterministic recurrent models can better match global temporal statistics under low entropy motion conditions.


This behavior is consistent with the conditional multi-modality of the datasets and the forecast horizons used in our study. In case of BAIR, the 
VRNN attains the best JEDi at longer horizons e.g., 5–20 and 15–15 in both reconstruction and transfer modes, indicating that latent variable dynamics better capture multi-modal future evolution as the prediction horizon increases. In contrast GRU is favored on MGIF/VoxCeleb where the motion is more repeatable or primarily involves subtle facial changes, and deterministic models are sufficient to match global temporal feature statistics.

\begin{figure*}[p]
    \centering
    \includegraphics[width=\textwidth]{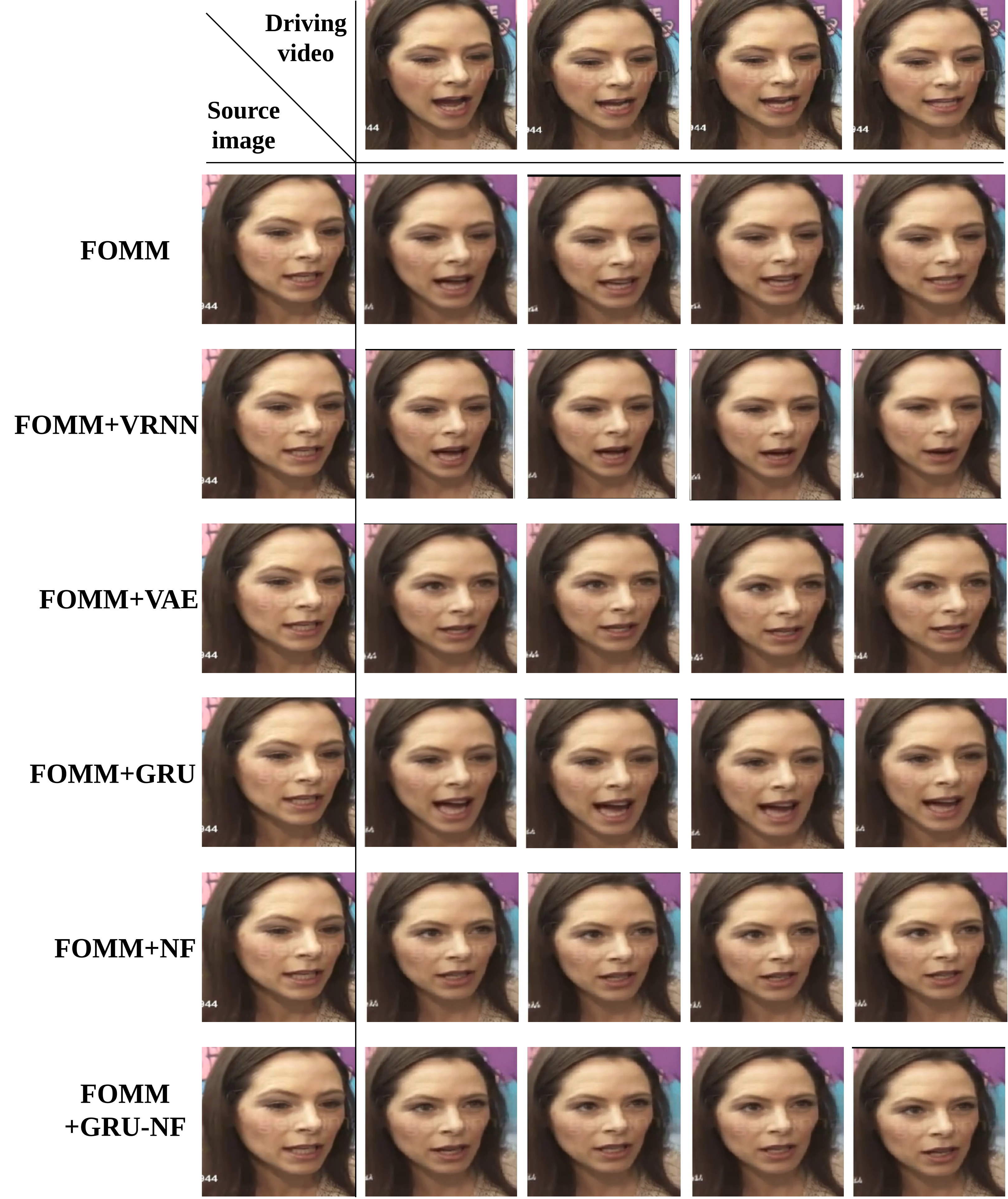}
    \caption{Qualitative results for VoxCeleb dataset in reconstruction mode}
    \label{fig:single prediction recon}
\end{figure*}

\begin{figure*}[p]
    \centering
    \includegraphics[width=\textwidth]{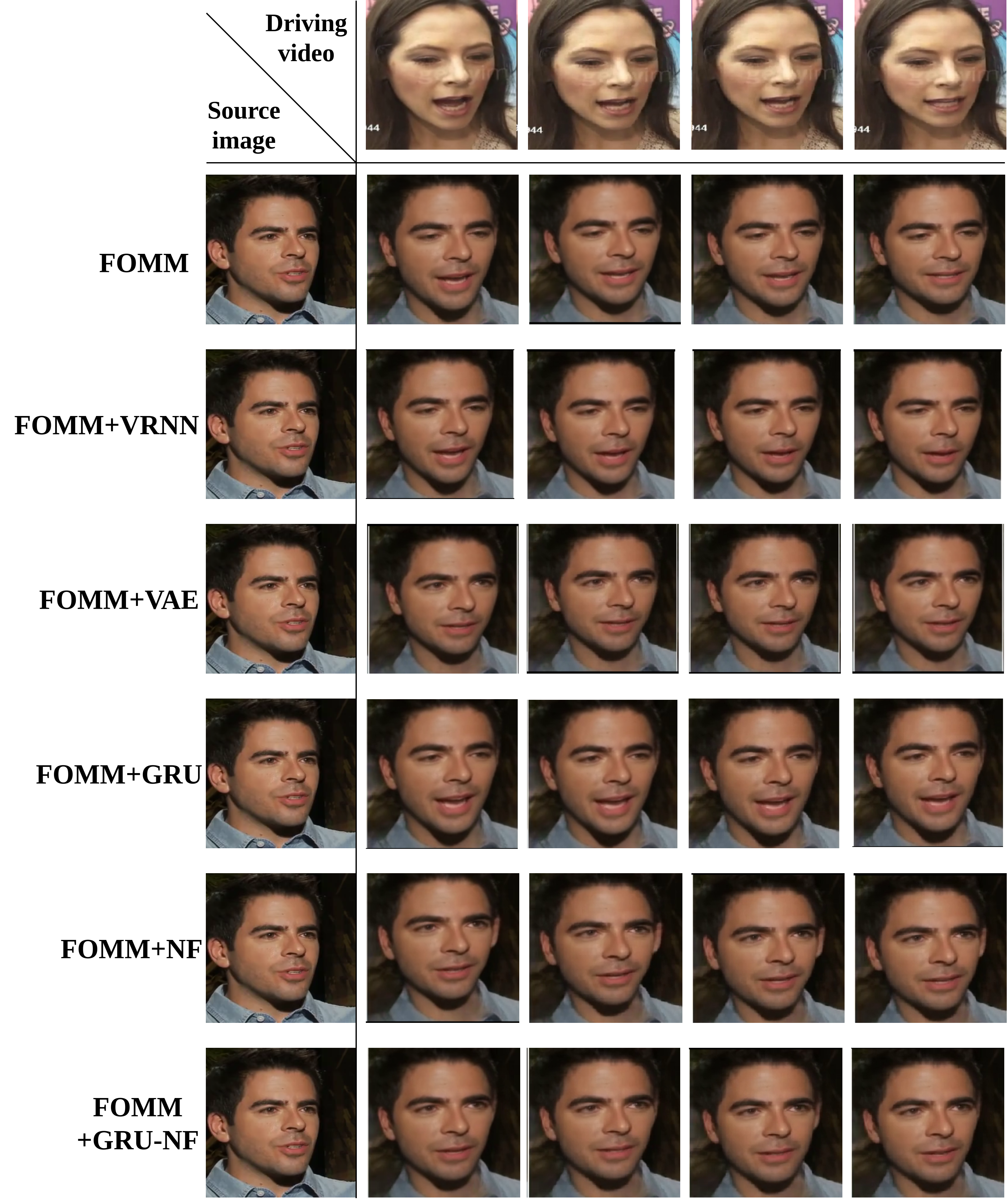}
    \caption{Qualitative results for VoxCeleb dataset in transfer mode}
    \label{fig:single prediction transfer}
\end{figure*}

Qualitative results using frames of generated videos are also shown for VoxCeleb data for both reconstruction and transfer modes and 12-12 prediction horizon in Figures \ref{fig:single prediction recon} and \ref{fig:single prediction transfer} respectively for single sample prediction. As reference the second row of the figure shows the video frames that are generated using the FOMM pipeline without keypoint prediction. For reconstruction mode, driving video sequence shown in the first row is the ground truth video and for transfer mode, FOMM generated video sequence is the ground truth. The third, fourth, fifth, sixth and seventh rows show the video frames generated using VRNN, VAE, GRU, NF and GRU-NF respectively for keypoint prediction and FOMM pipeline. 

\begin{table*}[t]
  \caption{MAE and JEDi results of MGIF dataset on reconstruction mode. For each criteria i.e. MAE and JEDi the best model is highlighted in bold, lower the better.}
  \label{table:1}
  \centering
  {\renewcommand{\arraystretch}{1.2}\setlength{\extrarowheight}{2pt}
  \resizebox{\textwidth}{!}{%
  \scalebox{1}[1.15]{%
  \begin{tabular}{l*{10}{c}}
    \toprule
        Prediction   & \multicolumn{5}{c}{MAE} & \multicolumn{5}{c}{JEDi} \\
    \cmidrule(lr){2-6} \cmidrule(lr){7-11}
    range
            & VRNN & VAE & GRU & NF & GRU-NF
            & VRNN & VAE & GRU & NF & GRU-NF \\
    \midrule
    6--6     &\textbf{0.0316} & 0.0392 & 0.0373 & 0.0379 & 0.0376
            & 6.3555 & 6.3117 & 6.4354 & 6.2996 & \textbf{6.2336} \\
    6--24     &0.0471	&0.0501	&\textbf{0.0451}	&0.0467	&0.0454
            & 6.5361	&7.4836	&\textbf{6.1183}	&6.4495	&6.7648 \\
    12--12   & \textbf{0.0375} & 0.0388 & 0.0379 & 0.0379 & 0.0377
            & \textbf{5.8972} & 6.5748 & 6.4593 & 6.0850 & 6.0207 \\
    \bottomrule
  \end{tabular}}}%
  }

  \caption{MAE and JEDi results of MGIF dataset on transfer mode. For each criteria i.e. MAE and JEDi the best model is highlighted in bold, lower the better.}
  \label{table:2}
  \centering
  {\renewcommand{\arraystretch}{1.2}\setlength{\extrarowheight}{2pt}
  \resizebox{\textwidth}{!}{%
  \scalebox{1}[1.15]{%
  \begin{tabular}{l*{10}{c}}
    \toprule
        Prediction   & \multicolumn{5}{c}{MAE} & \multicolumn{5}{c}{JEDi} \\
    \cmidrule(lr){2-6} \cmidrule(lr){7-11}
    range
            & VRNN & VAE & GRU & NF & GRU-NF
            & VRNN & VAE & GRU & NF & GRU-NF \\
    \midrule
    6--6     & \textbf{0.0192} & 0.0215 & 0.0195 & 0.0201 & 0.0197
            & 0.6635 & 0.9997 & \textbf{0.4312} & 1.0332 & 0.9396 \\
    6--24     & 0.0343	&0.0385	&\textbf{0.0319}	&0.0339	&0.0321
            & 0.8436	&5.2887	&\textbf{0.695}	&1.0901	&1.2615 \\
    12--12   &\textbf{0.0199} & 0.0215 & \textbf{0.0199} & 0.0202 & 0.0201
            & 0.983 & 0.464 & \textbf{0.423} & 1.2005 & 0.8449 \\
    \bottomrule
  \end{tabular}}}%
  }
\end{table*}

\begin{table*}[t]
  \caption{MAE and JEDi results of BAIR dataset on reconstruction mode. For each criteria i.e. MAE and JEDi the best model is highlighted in bold, lower the better.}
  \label{table:3}
  \centering
  {\renewcommand{\arraystretch}{1.2}\setlength{\extrarowheight}{2pt}
  \resizebox{\textwidth}{!}{%
  \scalebox{1}[1.15]{%
  \begin{tabular}{l*{10}{c}}
    \toprule
        Prediction    & \multicolumn{5}{c}{MAE} & \multicolumn{5}{c}{JEDi} \\
    \cmidrule(lr){2-6} \cmidrule(lr){7-11}
    range
            & VRNN & VAE & GRU & NF & GRU-NF
            & VRNN & VAE & GRU & NF & GRU-NF \\
    \midrule
    5--5     & \textbf{0.0713} &0.0749 &0.0722	&0.0766	&0.0763
            &1.655 &2.3094 &\textbf{1.6094} &2.5065 &2.7014 \\
    5--20     & \textbf{0.0891}	&0.0911	&0.09	&0.0911	&0.091
            & \textbf{6.2676}	&6.8537	&7.8202	&7.1344	&8.3909 \\
    15--15  & 0.0765 & \textbf{0.0759} &0.0773   &0.0793 &0.0791
            	&\textbf{3.2155} &3.7483 &3.9859 &3.4577 &3.665 \\
    \bottomrule
  \end{tabular}}}%
  }

  \caption{MAE and JEDi results of BAIR dataset on transfer mode. For each criteria i.e. MAE and JEDi the best model is highlighted in bold, lower the better.}
  \label{table:4}
  \centering
  {\renewcommand{\arraystretch}{1.2}\setlength{\extrarowheight}{2pt}
  \resizebox{\textwidth}{!}{%
  \scalebox{1}[1.15]{%
  \begin{tabular}{l*{10}{c}}
    \toprule
        Prediction    & \multicolumn{5}{c}{MAE} & \multicolumn{5}{c}{JEDi} \\
    \cmidrule(lr){2-6} \cmidrule(lr){7-11}
    range
            & VRNN & VAE & GRU & NF & GRU-NF
            & VRNN & VAE & GRU & NF & GRU-NF \\
    \midrule
    5--5     & \textbf{0.0199} &0.0267 &0.0216 &0.0272 &0.0267
            & 0.9274 &1.5357 &\textbf{0.5262} &1.5638 &1.9548 \\
    5--20    &\textbf{0.0449}	&0.0468	&0.0456	&0.0468	&0.0467
            & \textbf{5.7885}	&6.8148	&7.9668	&7.1783	&7.734 \\
    15--15  & \textbf{0.0258} &0.0263 &0.0277 &0.03 &0.0298
            	&\textbf{1.8371} &3.8864	&3.5842	&3.1205	&2.8107 \\
    \bottomrule
  \end{tabular}}}%
  }
\end{table*}

\begin{table*}[t]
  \caption{MAE and JEDi results of VoxCeleb dataset on reconstruction mode. For each criteria i.e. MAE and JEDi the best model is highlighted in bold, lower the better.}
  \label{table:5}
  \centering
  {\renewcommand{\arraystretch}{1.2}\setlength{\extrarowheight}{2pt}
  \resizebox{\textwidth}{!}{%
  \scalebox{1}[1.15]{%
  \begin{tabular}{l*{10}{c}}
    \toprule
        Prediction    & \multicolumn{5}{c}{MAE} & \multicolumn{5}{c}{JEDi} \\
    \cmidrule(lr){2-6} \cmidrule(lr){7-11}
    range
            & VRNN & VAE & GRU & NF & GRU-NF
            & VRNN & VAE & GRU & NF & GRU-NF \\
    \midrule
    6--6     & \textbf{0.0511} &0.0602 &0.0514 &0.0729 &0.0715
            & 0.7079 &0.8189 & \textbf{0.6936} &2.3025	&2.2377 \\
    6--24     & \textbf{0.0707}	&0.0771	&0.0732	&0.0875	&0.0878
            & 1.105	& \textbf{0.9118}	&0.9462	&3.0365	&3.3046 \\
    12--12  & \textbf{0.0564} &0.0651 &0.0566 &0.0724 &0.0715
            	&0.7664 &1.2794 & \textbf{0.7302} &1.4351 &1.6561 \\
    \bottomrule
  \end{tabular}}}%
  }

  \caption{MAE and JEDi results of VoxCeleb dataset on transfer mode. For each criteria i.e. MAE and JEDi the best model is highlighted in bold, lower the better.}
  \label{table:6}
  \centering
  {\renewcommand{\arraystretch}{1.2}\setlength{\extrarowheight}{2pt}
  \resizebox{\textwidth}{!}{%
  \scalebox{1}[1.15]{%
  \begin{tabular}{l*{10}{c}}
    \toprule
        Prediction    & \multicolumn{5}{c}{MAE} & \multicolumn{5}{c}{JEDi} \\
    \cmidrule(lr){2-6} \cmidrule(lr){7-11}
    range
            & VRNN & VAE & GRU & NF & GRU-NF
            & VRNN & VAE & GRU & NF & GRU-NF \\
    \midrule
    6--6     & \textbf{0.0137} &0.0244 &0.0148 &0.0364 &0.0348
            &0.5918 & \textbf{0.5583} &0.5808	&2.4871	&2.0271 \\
    6--24     & 0.0442	&0.0445	&\textbf{0.0411}	&0.0539	&0.0553
            &1.2844	& \textbf{0.8407}	&0.9875	&3.4093	&2.8105 \\
    12--12  & \textbf{0.0203} &0.0287 &0.0205 &0.035 &0.0348
            	&0.8267 &0.7610 & \textbf{0.6743}	&1.1633	&1.2116 \\
    \bottomrule
  \end{tabular}}}%
  }
\end{table*}

\subsection{Diverse Video Sample Prediction}

We have conducted our experiments using VRNN and GRU-NF for diverse sample prediction at longer prediction horizons, i.e. 6–24 and 12–12 for VoxCeleb, and 5–20 and 15–15 for BAIR. For each test video, we generated 100 samples in both reconstruction and transfer modes. We then select $C=20$ video samples from the $D=100$ generated samples that have the lowest MAE values compared to ground truth to demonstrate the diversity-fidelity trade-off.  Here the values C and D have been selected to compare the diverse video generation capabilities of the two models and in general they will vary based on the given application. To assess the performance of diverse sample generation, we have used APD to measure the pairwise distance among these 20 generated samples. The model that produces more diverse samples will have a higher APD value, indicating that the samples differ from one another significantly. Following this, the ratio of APD to MAE of each model is calculated to evaluate diversity vs. fidelity. 
A higher APD-to-MAE ratio for a given model indicates that it can produce more diverse samples with an acceptable difference w.r.t. the ground truth. To assess perceptual similarity of the generated videos w.r.t. the ground truth 
we have also measured the SSIM values of the generated video frames.


\begin{table*}[t]
  \caption{Diversity vs. fidelity results of VoxCeleb dataset on reconstruction mode. For APD to MAE ratio the best model is highlighted in bold, higher the better.}
  \label{table:7}
  \centering
  \resizebox{\textwidth}{!}{%
  \begin{tabular}{llccccc}
    \toprule
    Prediction & Model & APD & MAE & SSIM & \multicolumn{2}{c}{\textbf{APD to MAE ratio}} \\
    \cmidrule(lr){6-7}
    range  &  &  &  & & Mean & Median \\
    \midrule
    \multirow{2}{*}{12 - 12} & VRNN & 0.0074 & 0.062 & 0.6914 & 0.1142 & 0.0876 \\
    & GRU-NF & 0.0407 & 0.0817 & 0.6235 & \textbf{1.3452} & \textbf{1.3254} \\
    \midrule
    \multirow{2}{*}{6 - 24} & VRNN & 0.0079 & 0.0814 & 0.6026 & 0.0819 & 0.0749 \\
    & GRU-NF & 0.0608 & 0.0999 & 0.5441 & \textbf{1.6321} & \textbf{1.3126} \\
    \bottomrule
  \end{tabular}%
  }
\end{table*}

\begin{table*}[t]
  \caption{Diversity vs. fidelity results of VoxCeleb dataset on transfer mode. For APD to MAE ratio the best model is highlighted in bold, higher the better.}
  \label{table:8}
  \centering
  \resizebox{\textwidth}{!}{%
  \begin{tabular}{llccccc}
    \toprule
    Prediction & Model & APD & MAE & SSIM & \multicolumn{2}{c}{\textbf{APD to MAE ratio}} \\
    \cmidrule(lr){6-7}
    range  &  &  &  & & Mean & Median \\
    \midrule
    \multirow{2}{*}{12 - 12} & VRNN & 0.007 & 0.0334 & 0.8497 & 1.268 & 1.0229 \\
    & GRU-NF & 0.0336 & 0.0464 & 0.7861 & \textbf{5.4636} & \textbf{5.3986} \\
    \midrule
    \multirow{2}{*}{6 - 24} & VRNN & 0.0074 & 0.0548 & 0.7306 & 0.3199 & 0.3217 \\
    & GRU-NF & 0.05 & 0.0701 & 0.6655 & \textbf{3.9243} & \textbf{3.6426} \\
    \bottomrule
  \end{tabular}%
  }
\end{table*}

\begin{table*}[t]
  \caption{Diversity vs. fidelity results of BAIR dataset on reconstruction mode. For APD to MAE ratio the best model is highlighted in bold, higher the better.}
  \label{table:9}
  \centering
  \resizebox{\textwidth}{!}{%
  \begin{tabular}{llccccc}
    \toprule
    Prediction & Model & APD & MAE & SSIM & \multicolumn{2}{c}{\textbf{APD to MAE ratio}} \\
    \cmidrule(lr){6-7}
    range  &  &  &  & & Mean & Median \\
    \midrule
    \multirow{2}{*}{15 - 15} & VRNN & 0.0113 & 0.0834 & 0.7413 & 0.4445 & 0.3318 \\
    & GRU-NF & 0.0269 & 0.0853 & 0.7377 & \textbf{1.7144} & \textbf{1.369} \\
    \midrule
    \multirow{2}{*}{5 - 20} & VRNN & 0.0103 & 0.096 & 0.714 & 0.2831 & 0.2355 \\
    & GRU-NF & 0.0253 & 0.0971 & 0.7106 & \textbf{1.095} & \textbf{0.9684} \\
    \bottomrule
  \end{tabular}%
  }

  \caption{Diversity vs. fidelity results of BAIR dataset on transfer mode. For APD to MAE ratio the best model is highlighted in bold, higher the better.}
  \label{table:10}
  \centering
  \resizebox{\textwidth}{!}{%
  \begin{tabular}{llccccc}
    \toprule
    Prediction & Model & APD & MAE & SSIM & \multicolumn{2}{c}{\textbf{APD to MAE ratio}} \\
    \cmidrule(lr){6-7}
    range  &  &  &  & & Mean & Median \\
    \midrule
    \multirow{2}{*}{15 - 15} & VRNN & 0.011 & 0.0364 & 0.9065 & 0.519 & 0.3444 \\
    & GRU-NF & 0.0268 & 0.0393 & 0.8996 & \textbf{1.4679} & \textbf{1.4192} \\
    \midrule
    \multirow{2}{*}{5 - 20} & VRNN & 0.0099 & 0.053 & 0.8609 & 0.339 & 0.2606 \\
    & GRU-NF & 0.0247 & 0.0541 & 0.857 & \textbf{1.2689} & \textbf{1.1047} \\
    \bottomrule
  \end{tabular}%
  }
\end{table*}

The diversity vs. fidelity results for the VoxCeleb dataset in reconstruction mode and transfer modes are shown in Tables \ref{table:7} and \ref{table:8} respectively. The corresponding results for the BAIR dataset in reconstruction mode and transfer modes are shown in Tables \ref{table:9} and \ref{table:10} respectively. Details of the calculations used for MAE and APD values used in these tables are described below.

First we calculate mean MAE over the samples with the lowest 20 MAE values among the 100 generated samples for each test video. Then we compute the mean of these over all test videos. The APD values are calculated in a similar manner corresponding to the 20 generated videos that have the lowest MAE values for each test video.
Following this, the APD and MAE values are standardized using min-max standardization as follows:

\begin{equation}
APD_{\text{scaled}} = \frac{APD - APD_{\min}}{APD_{\max} - APD_{\min}}
\end{equation}

\begin{equation}
MAE_{\text{scaled}} = \frac{MAE - MAE_{\min}}{MAE_{\max} - MAE_{\min}}
\end{equation}

\noindent Here, \(APD_{\min}\) and \(APD_{\max}\), and \(MAE_{\min}\) and \(MAE_{\max}\) represent the minimum and maximum values of the APD and MAE values respectively. 
These are determined using the complete set of APD and MAE values across all test videos and both models
to ensure a consistent normalization scale 
facilitating direct comparisons. After standardizing the APD and MAE values, the ratio of standardized APD to standardized MAE is calculated for each test video. The mean and median of these ratio values across all test videos are then reported in Tables \ref{table:7}, \ref{table:8}, \ref{table:9} and \ref{table:10}. Since APD and MAE represent different performance metrics, our standardization procedure ensures that their values remain within the same range, thereby allowing for a meaningful comparison of their ratios.

In Table \ref{table:7}, diverse sample prediction performance comparisons of VRNN and GRU-NF models are given for VoxCeleb dataset in reconstruction mode for 6-24 and 12-12 prediction range. It shows that in 12-12 prediction horizon GRU-NF produces more diverse samples compared to VRNN since the APD value of GRU-NF is more than 5.5 times higher than that of VRNN. Although the MAE value of VRNN is lower than that of GRU-NF, the APD to MAE ratio of GRU-NF indicates that the generated diverse samples in this case do not significantly deviate from the ground truth. The mean value of the APD to MAE ratios for GRU-NF is 11.78 times higher than that of VRNN, indicating that although the samples generated from VRNN’s latent space are close to the ground truth, they lack significant diversity. The median value of the ratios also reflects this trend, with a higher value for GRU-NF compared to VRNN. This suggests that, even across different test videos, GRU-NF consistently produces a diverse set of predictions with an acceptable deviation from ground truth. A higher median value observed in GRU-NF further reinforces that the higher APD to MAE ratio relative to VRNN is not due to a few outliers 
but a general characteristic of the model across the dataset. This trend can also be observed for the 6-24 prediction horizon. 
A similar pattern is seen in the results as given in Tables \ref{table:8}, \ref{table:9} and \ref{table:10}. In each case, the mean and median values of APD to MAE ratios are higher for GRU-NF than VRNN.

\begin{figure}[H]
\centering
\begin{minipage}{0.48\textwidth}
    \centering
    \includegraphics[width=\textwidth]{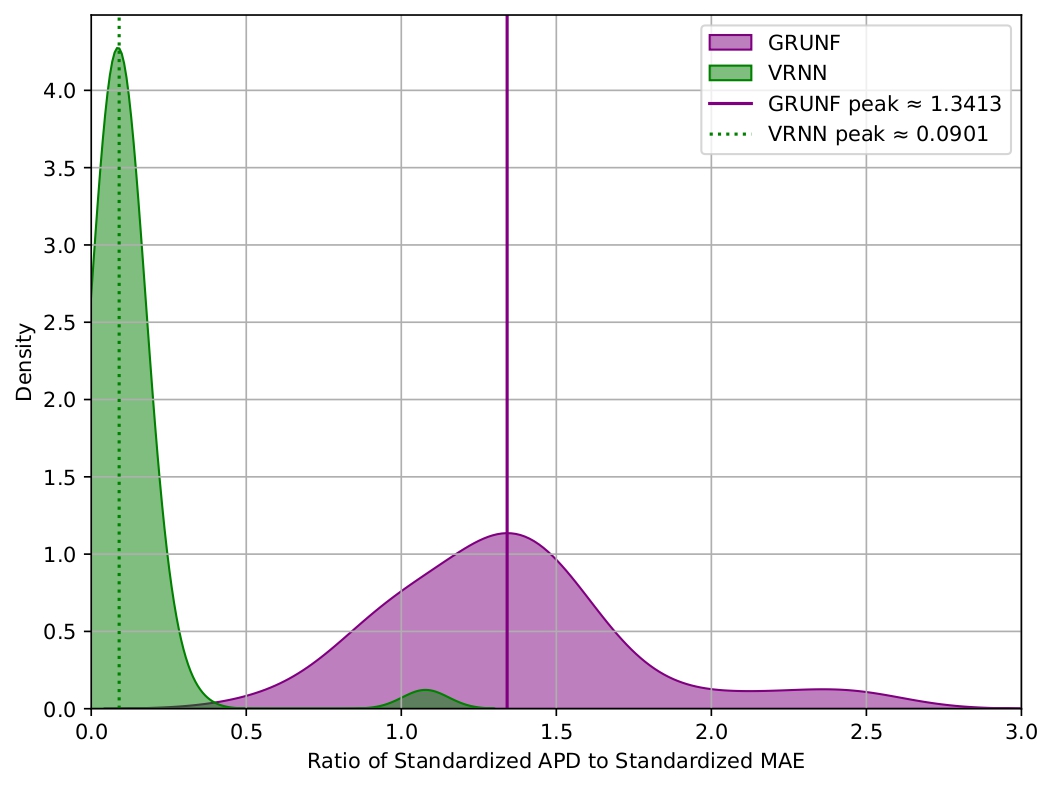}
    \captionof{figure}{Comparison of APD to MAE ratio distributions for VoxCeleb dataset in reconstruction mode}
    \label{fig:APD to MAE Ratio for vox recon}
\end{minipage}
\hfill
\begin{minipage}{0.48\textwidth}
    \centering
    \includegraphics[width=\textwidth]{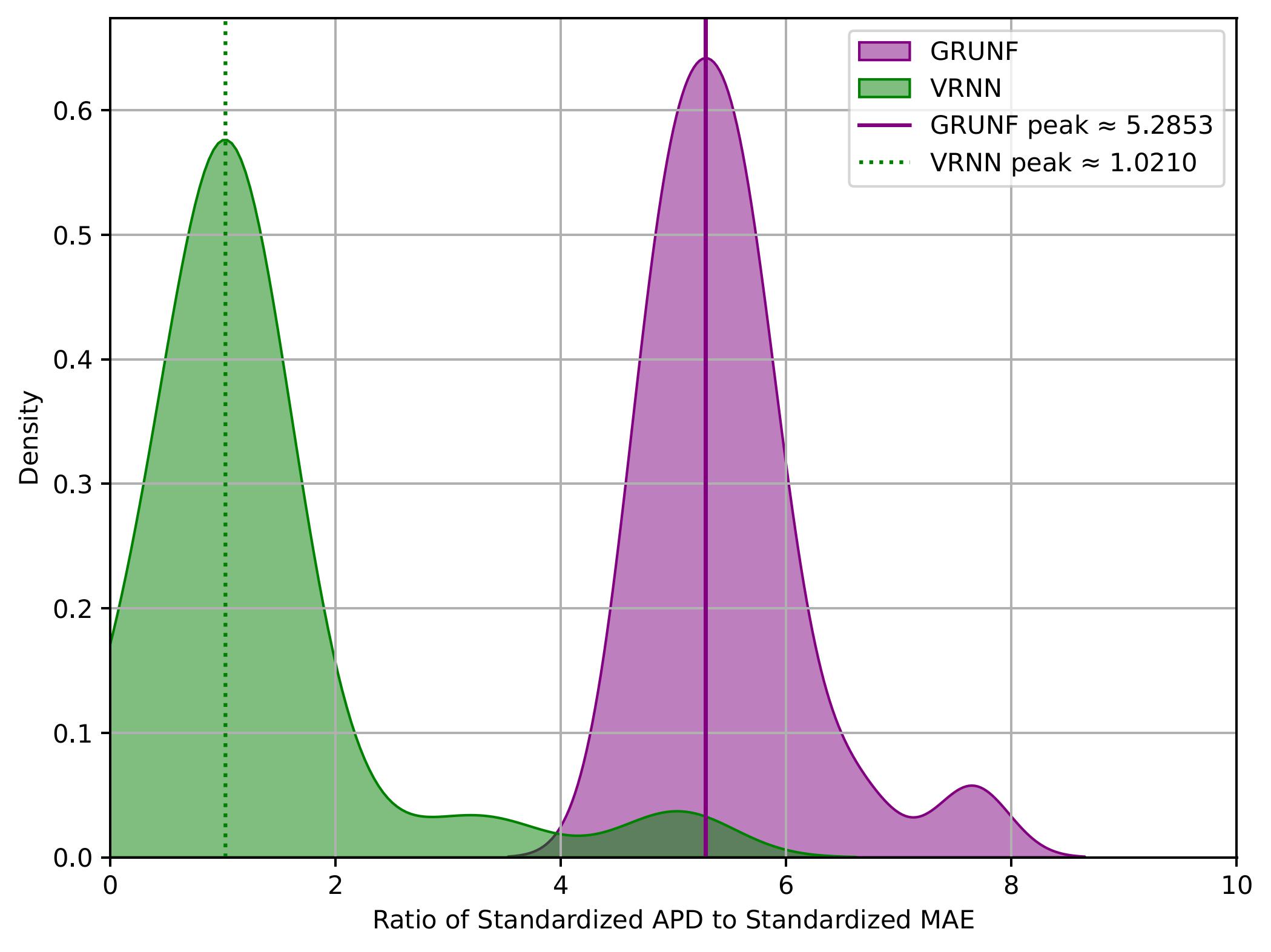}
    \captionof{figure}{Comparison of APD to MAE ratio distributions for VoxCeleb dataset in transfer mode}
    \label{fig:APD to MAE Ratio for vox transfer}
\end{minipage}

\vspace{0.4cm} 

\begin{minipage}{0.48\textwidth}
    \centering
    \includegraphics[width=\textwidth]{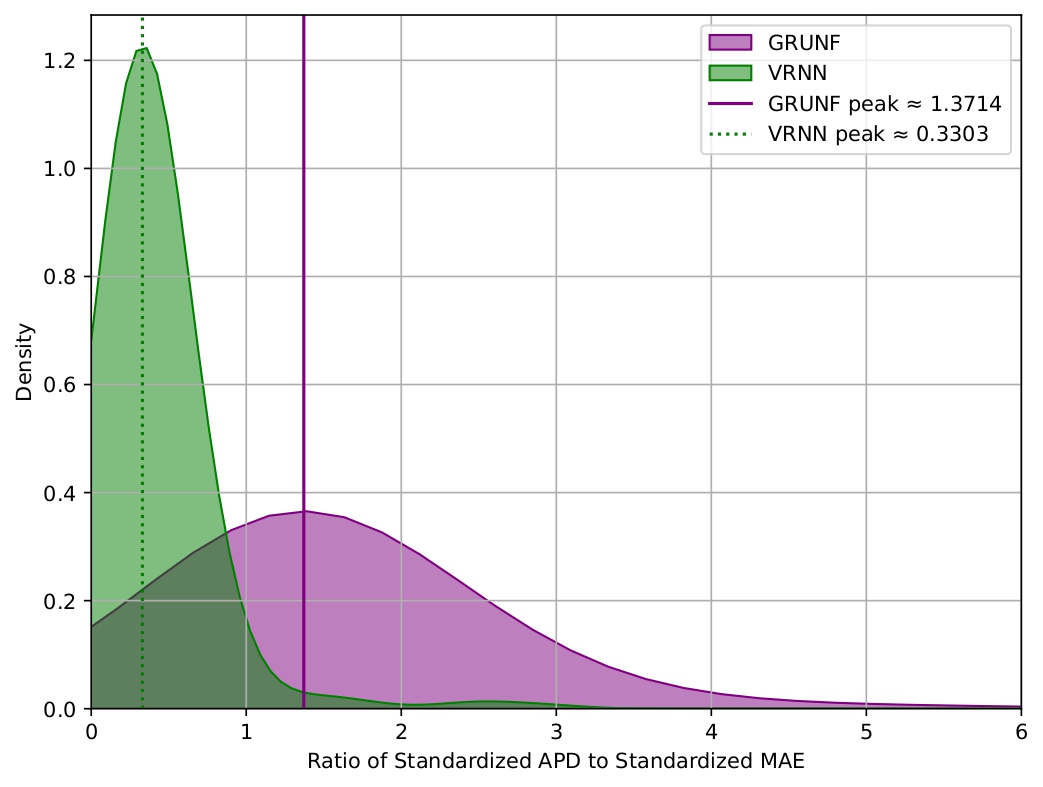}
    \captionof{figure}{Comparison of APD to MAE ratio distributions for BAIR dataset in reconstruction mode}
    \label{fig:APD to MAE Ratio for bair recon}
\end{minipage}
\hfill
\begin{minipage}{0.48\textwidth}
    \centering
    \includegraphics[width=\textwidth]{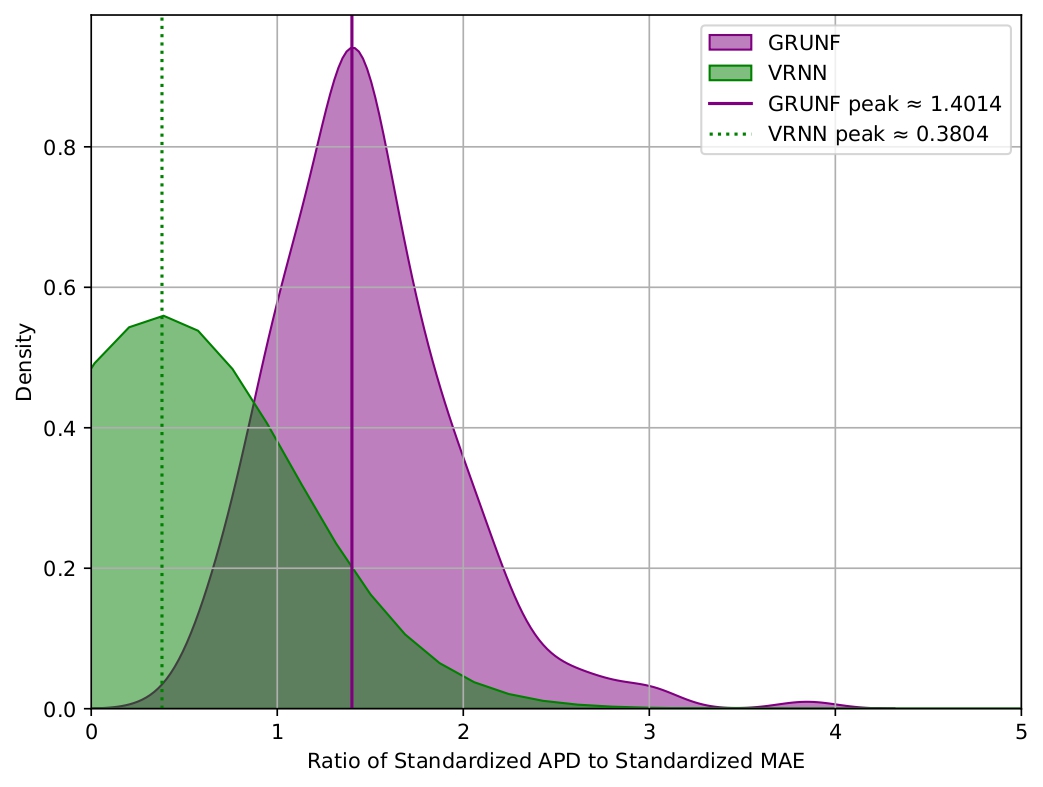}
    \captionof{figure}{Comparison of APD to MAE ratio distributions for BAIR dataset in transfer mode}
    \label{fig:APD to MAE Ratio for bair transfer}
\end{minipage}

\end{figure}

To provide a comparison of the standardized APD to standardized MAE ratio distributions for both models, we use kernel density plots shown in Figures \ref{fig:APD to MAE Ratio for vox recon} and \ref{fig:APD to MAE Ratio for vox transfer} for VoxCeleb dataset in reconstruction mode and transfer mode respectively for 12-12 prediction horizon. Similarly, the comparison for BAIR dataset in both modes for both models for 15-15 prediction horizon are shown in Figures \ref{fig:APD to MAE Ratio for bair recon} and \ref{fig:APD to MAE Ratio for bair transfer} for 15-15 prediction horizon. To summarize, across both datasets and both modes, the pattern remains consistent: the density peak of GRU-NF appears to the right of VRNN’s peak, meaning it prioritizes diversity but not at the cost of excessive error. Conversely, the density distributions of VRNN reflect slightly lower error along with lower diversity. Therefore, GRU-NF offers a better and reasonable trade-off between fidelity and diversity.

To assess the perceptual quality of the generated samples, we also perform a qualitative evaluation based on structural similarity. Specifically, for each test video, we select the 20 generated samples with the lowest MAE values and compute the SSIM between each of these samples and the corresponding ground truth. The mean of the SSIM values are then taken to obtain a single SSIM score per video. Finally, the mean SSIM value across all test videos is reported in the Tables \ref{table:7}, \ref{table:8}, \ref{table:9} and \ref{table:10}. The results indicate that the SSIM value is greater for VRNN than GRU-NF since the generated samples using VRNN have little deviation from ground truth compared to GRU-NF. 
However though VRNN has greater SSIM values than GRU-NF, SSIM values for GRU-NF fall in fair, good or excellent quality range \cite{Rehman2015}. Thus, the results of the APD to MAE ratio and SSIM value across both modes of both datasets demonstrate that the generated samples using GRU-NF are more diverse compared to VRNN, while having meaningful representations.

\begin{figure}[t]
  \centering
  \begin{subfigure}[b]{\textwidth}
    \includegraphics[width=0.8\textwidth, height=0.4\textheight]{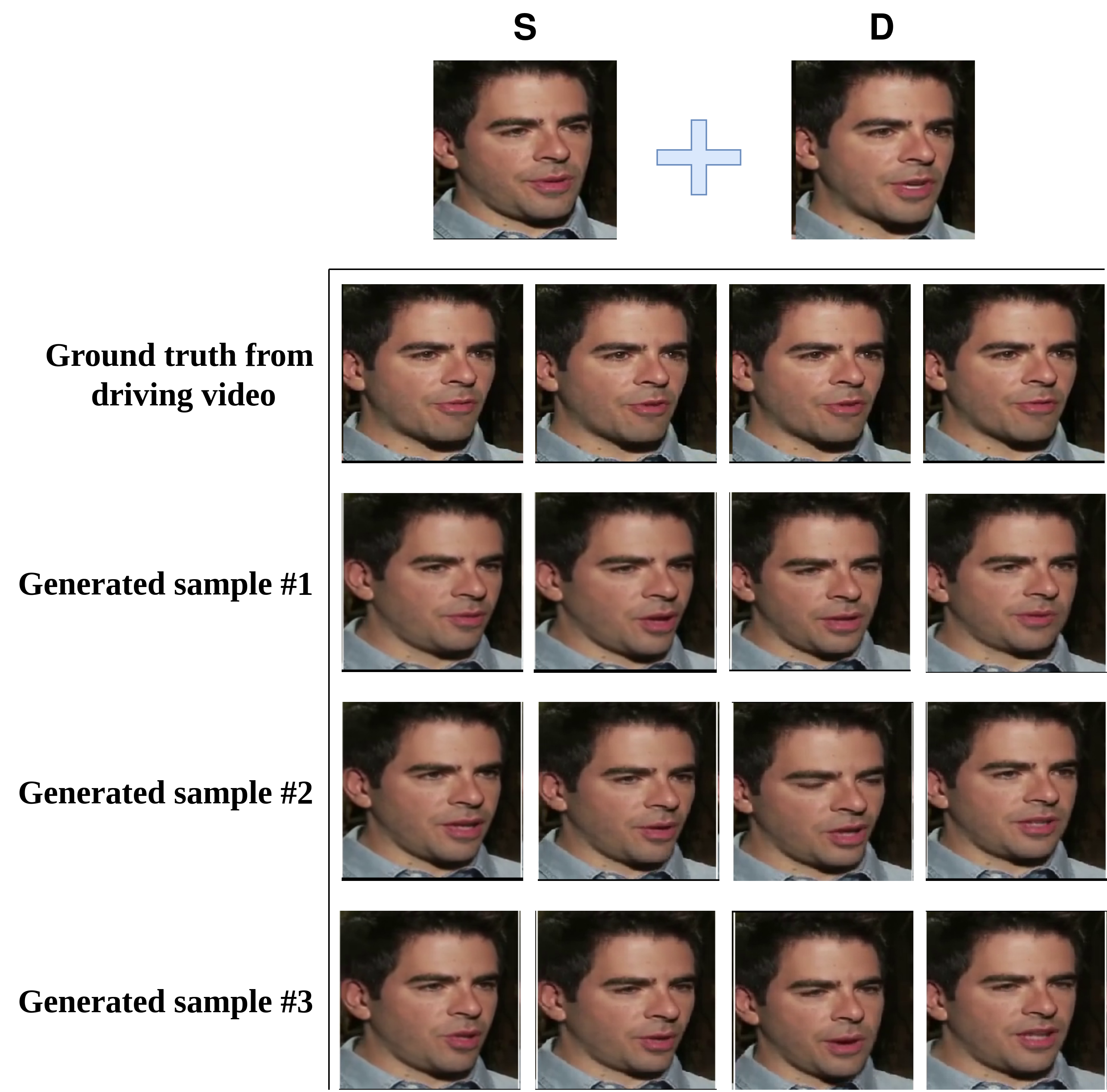}
    \caption{}
    \label{fig:vox_grunf_recon_diverse}
  \end{subfigure}
  \vspace{0.5cm}
  \begin{subfigure}[b]{\textwidth}
    \includegraphics[width=0.8\textwidth, height=0.4\textheight]{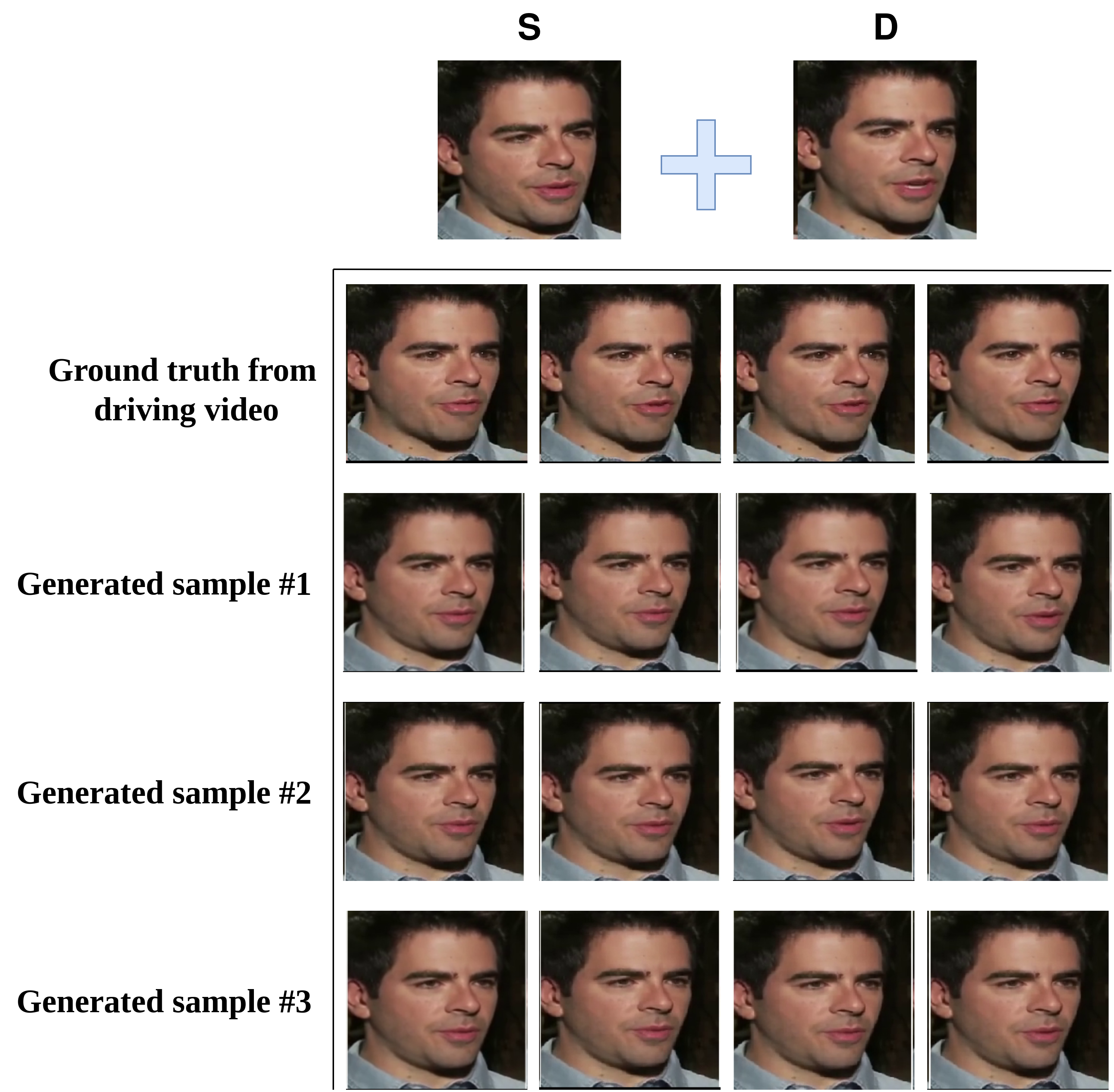}
    \caption{}
    \label{fig:vox_vrnn_recon_diverse}
  \end{subfigure}
  \caption{Qualitative results for VoxCeleb dataset in reconstruction mode in generating diverse samples using \textbf{(a)} GRU-NF and \textbf{(b)} VRNN for keypoint prediction}
  \label{fig:vox_recon_diverse}
\end{figure}

\begin{figure}[t]
  \centering
  \begin{subfigure}[b]{\textwidth}
    \includegraphics[width=0.8\textwidth, height=0.4\textheight]{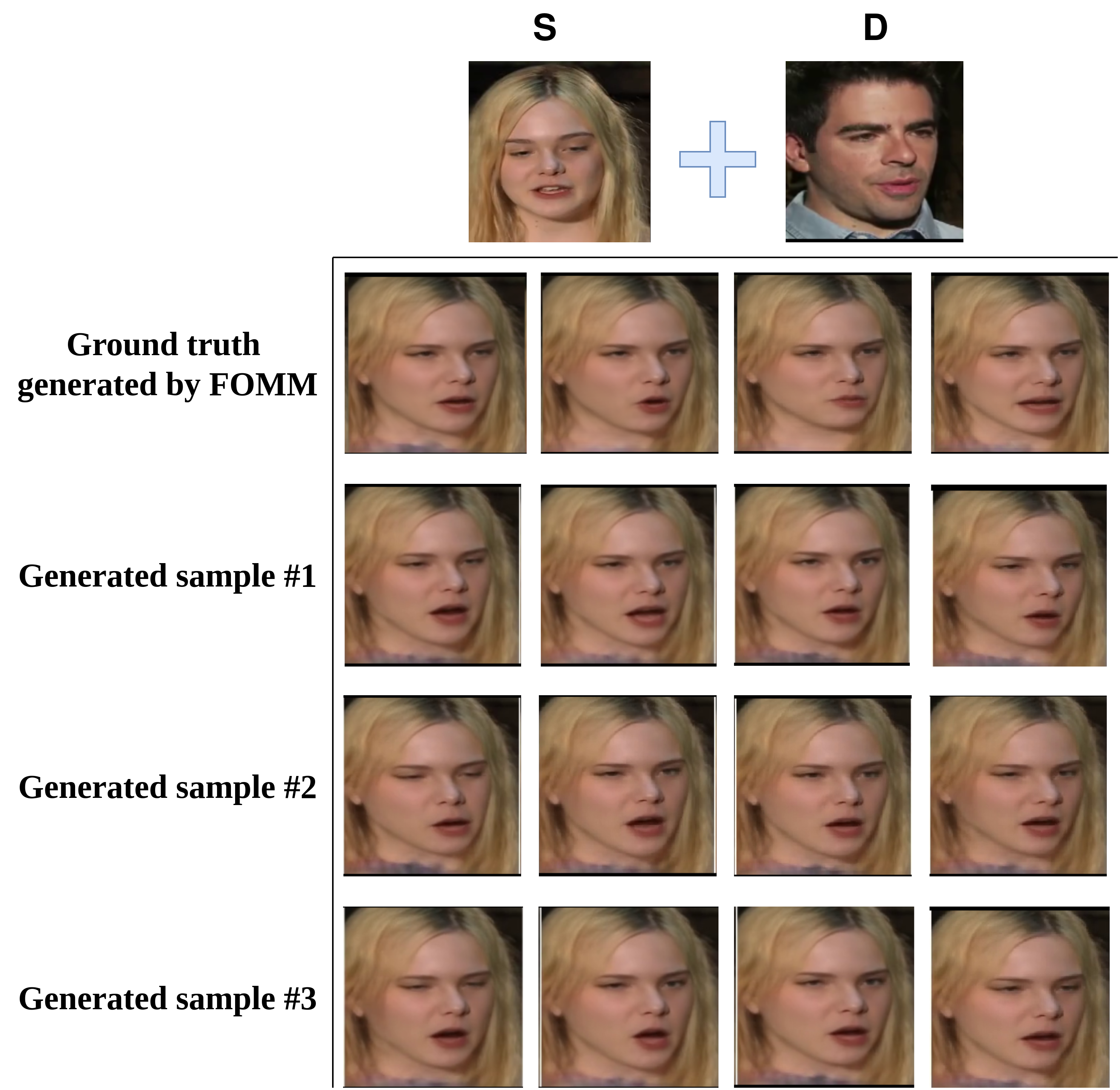}
    \caption{}
    \label{fig:vox_grunf_transfer_diverse}
  \end{subfigure}
  \vspace{0.5cm}
  \begin{subfigure}[b]{\textwidth}
    \includegraphics[width=0.8\textwidth, height=0.4\textheight]{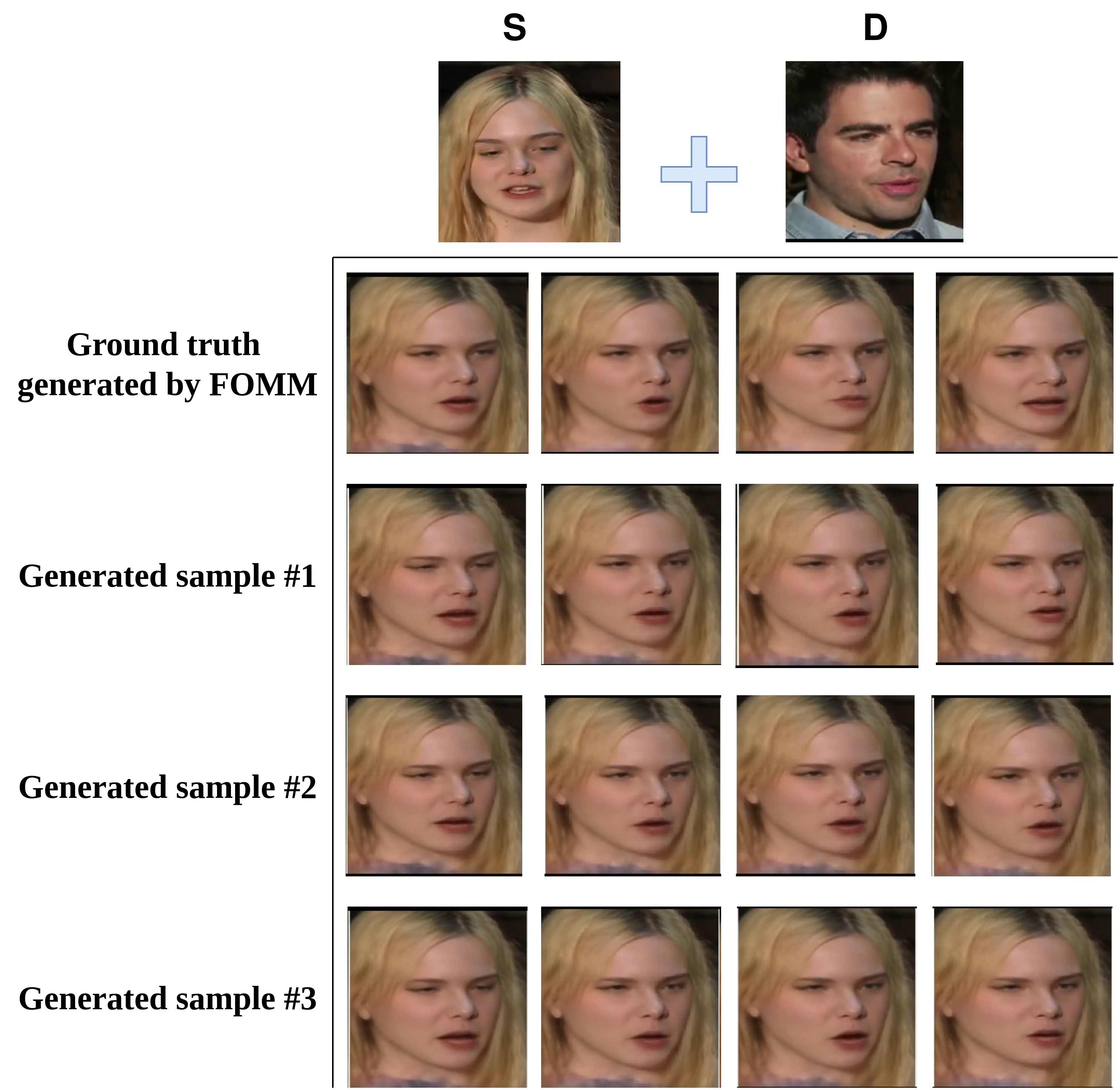}
    \caption{}
    \label{fig:vox_vrnn_transfer_diverse}
  \end{subfigure}
  \caption{Qualitative results for VoxCeleb dataset in transfer mode in generating diverse samples using \textbf{(a)} GRU-NF and \textbf{(b)} VRNN for keypoint prediction}
  \label{fig:vox_transfer_diverse}
\end{figure}

Qualitative results for diverse sample prediction are shown in Figures \ref{fig:vox_recon_diverse} and \ref{fig:vox_transfer_diverse} for VoxCeleb dataset for 12-12 prediction horizon. The Figures \ref{fig:vox_grunf_recon_diverse} and \ref{fig:vox_vrnn_recon_diverse} show the produced diverse samples in reconstruction mode using GRU-NF and VRNN for keypoint prediction respectively, and Figures \ref{fig:vox_grunf_transfer_diverse} and \ref{fig:vox_vrnn_transfer_diverse} show the produced diverse samples in transfer mode using GRU-NF and VRNN for keypoint prediction respectively. In each figure, 
the first row serves as the ground truth video sequence. The second, third and fourth rows show the examples of generated diverse video frames using the prediction model. It is evident from the figures that the facial expressions in the generated video samples using VRNN for keypoint prediction are nearly identical. In contrast, the generated samples using GRU-NF for keypoint prediction exhibit distinct facial expressions, demonstrating greater variation while remaining meaningful.

\newpage

\subsection{Throughput, Compute and Bandwidth for Real-Time Use}

\noindent \textbf{Setup.} We measure end-to-end latency of the full pipeline of our model: keypoint extraction → forecasting → video generation for single video sample prediction, on an NVIDIA RTX 3090 (24 GB VRAM), CUDA 12.x, PyTorch FP32 (Full precision 32 bit, no quantization) using 256×256 video frames from the VoxCeleb dataset. We compare this  versus other keypoint-based motion transfer models; namely the Thin Plate Spline (TPS) model integrated with VRNN and the Neural Talking Head model integrated with VRNN by measuring latency for them on the same hardware. 

\textbf{Inference time and fps estimate.} For each model we process 36 consecutive frames arranged as blocks of 6 observed frames and 6 predicted frames in a repeating pattern. Thus, over 36 frames we have 18 observed (used for keypoint extraction) and 18 predicted (produced by the VRNN), while video generation runs on all 36 frames. We record per-stage wall-clock time and aggregate these to obtain the total inference time for 36 frames. To report performance 
as per standard videoconferencing setups, we use linear scaling to obtain the inference time T for 30 frames; following which the frames per second (fps) is estimated as 30/T.

\textbf{TOPS estimate.} Our steady-state compute accounting follows stride-2 operation across the 3 stages of the pipeline: computations during  keypoint extraction and  keypoint forecasting occur only during half of the total 30 frames, whereas video generation runs on every frame as mentioned earlier. 
We incorporate these duty cycle considerations when converting per-stage operation counts to model-required Trillion
Operations per second (TOPS). Operation counts are obtained using fvcore’s FlopCountAnalysis on representative inputs; fvcore traces the model and reports a per-module flop (MAC) estimate \cite{fvcore}. We then report  TOPS = (FLOPs per output frame × fps) / (u × 10¹²) assuming an average device utilization of 60\% \cite{gao2024empirical}, i.e. u=0.6. These differences in compute across the different stages are taken into account during our TOPS calculations and the reasoning generalizes to other strides. 

\textbf{Bitrate estimate.} For keypoint based motion transfer models, the bitrate estimate $R_{model}$  for keypoints we use FP32 data representation format with $R_{model}$  (kbps)=D×32×30/1000=0.96D at 30 fps; where D is the keypoint dimension. Using D=60 (FOMM), D=100 (TPS), and D=51 (Neural Talking Head), these rates are 57.6, 96.0, and 48.96 kbps, respectively. With forecasting of half of the total number of frames (15 out of 30), the rate halves to 0.48D. Savings are computed versus a practical ~500 kbps baseline for a frame size of 256×256 at 30 fps  {standard video-conferencing streams without keypoint extraction and keypoint forecasting} (as obtained from publicly available specifications) \cite{zoom_system_requirements_desktop, webex_min_bandwidth_meetings}. For each model, the bandwidth saving is then calculated using 500/$R_{model}$.



\begin{table}[ht]
\centering
\caption{Comparison of inference time for the entire pipeline with other motion transfer models. For each criteria i.e. total inference time and total compute the best model is highlighted in bold and the second best is highlighted by underline, lower the better. Measurements were taken on 36 frames and linearly scaled to 30 frames.}
\label{tab:pipeline-inference}
\footnotesize
\setlength{\tabcolsep}{10pt}
\renewcommand{\arraystretch}{1.06}
\setlength{\extrarowheight}{2pt}
\begin{tabularx}{\linewidth}{@{} l c c cc cc cc @{}}
\toprule
\multirow{2}{*}{Model} &
\multirow{2}{*}{\makecell{Total\\time (s)}} &
\multirow{2}{*}{\makecell{Total\\TOPS@30 fps}} &
\multicolumn{2}{c}{Keypoint extraction} &
\multicolumn{2}{c}{Keypoint forecast} &
\multicolumn{2}{c}{Video generation} \\
\cmidrule(lr){4-5}\cmidrule(lr){6-7}\cmidrule(lr){8-9}
& & &
\makecell{time (s)} & \makecell{TOPS} &
\makecell{time (s)} & \makecell{TOPS} &
\makecell{time (s)} & \makecell{TOPS} \\
\midrule
FOMM+VRNN                  & \textbf{0.8176}  & \textbf{2.7194}  & 0.0336 & 0.0310 & 0.0743 & 0.0001 & 0.7097 & 2.6883 \\
TPS+VRNN                   & \underline{1.0627} & \underline{6.4210} & 0.0283 & 0.0592 & 0.0878 & 0.0001 & 0.9466 & 6.3617 \\
\makecell[l]{Neural Talking\\Head+VRNN} & 1.2999 & 10.0026 & 0.0893 & 0.1958 & 0.0782 & 0.0001 & 1.1323 & 9.8067 \\
\bottomrule
\end{tabularx}

\centering
\caption{Comparison of throughput, compute, bandwidth savings and accuracy with other motion transfer models. For each column the best model is highlighted in bold and the second best is highlighted by underline. For throughput and bandwidth saving, higher is better; for compute, MAE and JEDi, lower is better.}
\label{tab:pipeline-compute}
\setlength{\tabcolsep}{15pt}
\renewcommand{\arraystretch}{1.12}
\setlength{\extrarowheight}{2pt}
\begin{tabularx}{\linewidth}{@{} l c c c c c @{}}
\toprule
Model &
\makecell{Throughput\\(fps)} &
\makecell{Compute\\(TOPS)} &
\makecell{Bandwidth\\Saving ($\times$)} &
MAE & JEDi \\
\midrule
FOMM+VRNN                & \textbf{36.69} & \textbf{2.7194} & \underline{17.36} & \underline{0.0511} & \underline{0.7079} \\
TPS+VRNN                 & \underline{28.23} & \underline{6.4210} & 10.40 & \textbf{0.0499} & \textbf{0.5716} \\
Neural Talking Head+VRNN & 23.08 & 10.0026 & \textbf{20.42} & 0.0619 & 0.738 \\
\bottomrule
\end{tabularx}
\end{table}

We benchmark all 3 models on VoxCeleb dataset in reconstruction mode using a 6-6 prediction horizon. The inference time and compute per-stage (keypoint extraction, keypoint forecast and video generation) to generate 30 consecutive frames are displayed in Table \ref{tab:pipeline-inference} and the throughput, total compute (TOPS) for the entire pipeline, bandwidth saving, MAE and JEDi are displayed in Table \ref{tab:pipeline-compute} for all three models. From the results we can observe that, FOMM+VRNN is the most suitable choice for real-time, resource-constrained deployment: it takes the lowest time (0.8176 s) to generate consecutive 30 frames which is 23.07\% lower than the time required for the second best model TPS+VRNN. Moreover, it is the fastest at 36.69 fps and needs the least compute (2.7194 TOPS) which is 29.97\% faster and 57.65\% lesser than TPS+VRNN, respectively. The lowest bitrate estimate is 24.48 kbps for Neural Talking Head+VRNN models which provides the highest bandwidth saving 20.42x vs a 500 kbps baseline, however it attains the worst MAE and JEDi among the competing models. FOMM+VRNN model requires 28.8 kbps which leads to 17.36× saving, providing near-best bandwidth reduction while also achieving near-best MAE and JEDi. Therefore, the substantial gains in fps, compute, and bandwidth make FOMM+VRNN the best overall real-time option.

\textbf{Selection rationale.} For real-time, resource-constrained deployment (edge or interactive), FOMM+VRNN offers the best Pareto balance: it is the fastest (36.69 fps), requires the least compute (2.7194 TOPS), and achieves near-best bandwidth reduction, while keeping MAE within \(\approx 2\%\)
 of the most accurate variant. TPS+VRNN remains a quality-first alternative when maximal perceptual accuracy (MAE/JEDi) is prioritized and runtime/compute are less constrained.

 \section{Discussion}

The algorithms used for single sample and diverse sample video prediction exhibit different characteristics, making them suitable for different applications based on specific requirements. A comparative description is provided below.

\subsection{Single Video Sample Prediction}


Among the evaluated models (VRNN, VAE, NF, GRU, and GRU-NF), the results indicate that performance depends on the dataset regime and the prediction horizon, rather than one model dominating uniformly across all settings. This behavior aligns with the different inductive biases of the models and with how conditional uncertainty evolves with horizon. VRNN is particularly well-suited for modeling sequential data that exhibits complex temporal structure, high intrinsic variability, and a high signal-to-noise ratio. It explicitly introduces time-dependent latent variables coupled with recurrence, enabling it to represent latent stochastic factors that influence future motion and become increasingly important as the forecast horizon grows. Consequently, VRNN shows clearer advantages in settings that are more conditionally multi-modal i.e., multiple plausible futures from similar prefixes, which is most evident on BAIR at longer horizons. In contrast, GRU is deterministic and often provides stable rollouts when the conditional dynamics are closer to unimodal e.g., phase-locked or low-amplitude motion. This helps explain why GRU shows superior performance in MGIF and VoxCeleb datasets, particularly at longer horizons where small temporal inconsistencies can accumulate.

VAEs, while effective at learning probabilistic latent representations and generating diverse samples, are not inherently designed to model temporal dependencies, and thus may underperform in capturing the evolving structure of keypoint motion. NF models, despite offering exact likelihood computation and expressiveness in latent space, can struggle to model temporal coherence when applied directly to sequences, as they lack built-in mechanisms to capture sequence dependencies. The hybrid GRU-NF, although combining temporal modeling from GRU with expressiveness from NF, still underperforms compared to VRNN. Since two parts (GRU and NF) of the model aren't learning collaboratively, which means the GRU isn’t updated based on the latent sample generated by NF, its hidden states may not provide the most useful information for the NF especially in the presence of complex, high-variation motion patterns. Similarly, the NF may focus on fitting the data distribution well, but without proper feedback from the temporal learning component, it might not fully leverage the temporal structure encoded by the GRU. This contrasts with VRNN, where the recurrent structure and the latent variables are trained end-to-end in a tightly coupled fashion, along with minimizing reconstruction loss ensuring that the model learns both temporal dynamics and stochastic variation in a coordinated way.

\subsection{Diverse Video Sample Prediction}

Except the GRU all the models studied in this work have a latent space that helps them to produce more than one plausible outcome. 
Although VAE and NF are able to produce diverse samples, they lack the ability to capture temporal diversity, limiting their effectiveness in modeling dynamic, time-evolving structures. Therefore, we have excluded these models when considering diverse video sample prediction.

The results from our simulations using both VRNN and GRU-NF show that each model utilizes its latent space uniquely. Consequently, VRNN demonstrates better performance in terms of MAE, whereas GRU-NF provides better diversity among samples. This distinction suggests different potential applications for each model.

VRNN is a VAE based architecture which relies on approximate likelihood estimation, that can hinder its ability to capture the full complexity of the underlying data distribution. 
Since GRU-NF is based on NF architecture that uses exact likelihood estimation, it learns a more expressive latent space with flexible transformations, leading to a wider and more complex distribution. The substantial differences between the latent samples lead to a comparatively higher deviation from ground truth at the cost of a wider variety of predictions. This trade-off highlights the strength of exact likelihood methods like NF in capturing the complete probability distribution, which is especially valuable for applications requiring probabilistic forecasting or uncertainty quantification.

\section{Conclusions and Future Work}

In this research, we aim to investigate the strengths and limitations of 
generative time series models for both single and diverse sample prediction in video datasets. We explore a combination of the GRU
which is a sequence prediction model with two different generative models, VAE and NF. The first combination, known as VRNN, pairs GRU with VAE, while the second, GRU-NF, integrates GRU with Normalizing Flows.  

Our findings indicate that in single video sample prediction, VRNN achieves the best point-forecast fidelity (lowest MAE) and can potentially enable
2x additional bandwidth reduction in existing motion transfer pipelines. It is particularly competitive in datasets that are more ambiguous and multi-modal. In contrast, GRU attains the best performance in settings where the motion is more repeatable or involves subtle changes, thereby suggesting that stable deterministic models can align well with global temporal feature statistics. These findings highlight that the choice of architecture should be guided by the conditional uncertainty of the dataset.
In the case of diverse video sample prediction, GRU-NF proves superior in terms of the diversity-fidelity trade-off. While VRNN generates samples closely resembling the ground truth with minimal variation among them, GRU-NF produces more diverse samples without significantly compromising visual quality. This highlights the fundamental distinction between the two approaches: VRNN prioritizes accuracy at the expense of diversity, whereas GRU-NF does the opposite. 

The choice between these models depends on the specific application for real-time motion transfer depending on whether the task requires precision or diversity.
In scenarios such as video conferencing and remote patient monitoring, where accurate forecasting is critical, VRNN or GRU is preferred depending on dataset characteristics: VRNN is more suitable for higher multi-modality settings e.g., posture changes or irregular patient movement, while GRU can be competitive for more constrained or periodic motion e.g., smooth head movements. Conversely, applications like VR gaming and vision-based anomaly detection in the manufacturing industry require diverse sample generation at future timesteps with high fidelity, making GRU-NF the more suitable option. 
Our future work will focus on investigation of these trade-offs involved in using various generative time series models for different real-time motion transfer applications on resource constrained edge computing platforms. 
In addition we plan to enhance FOMM’s lightweight motion transfer pipeline by considering application driven upgrades to its keypoint based motion modeling. This will be done by incorporating more expressive geometric transforms to maintain the required semantic and spatial consistency while explicitly preserving the superior real-time characteristics of FOMM and ease of integration with generative time series models.

\section*{Statements and Declarations}

\subsection*{Conflict of Interest Statement}

The authors declare that the research was conducted in the absence of any commercial or financial relationships that could be construed as a potential conflict of interest.

\subsection*{Author Contributions}

TH: Methodology, Investigation, Validation, Visualization, Writing – original draft. MABS: Validation, Visualization, Writing –original draft. BJ: Software. XB: Software. SM: Writing – review \& editing, SP: Writing – review \& editing, IA: Writing – review \& editing, SD: Conceptualization, Methodology, Resources, Supervision, Writing – review \& editing.

\subsection*{Funding}
The author(s) declare that financial support was received for the research of this article. Tasmiah Haque was partially supported for this work by an Intel grant. 

\subsection*{Acknowledgments}
The authors would like to acknowledge the Pacific Research Platform, NSF Project ACI-1541349, and Larry Smarr (PI, Calit2 at UCSD) for providing the computing infrastructure used in this project. The authors also thank Md Mushfiqur Rahaman for helpful technical discussions and computing support.

\subsection*{Data Availability Statement}
The datasets used to conduct this study are available from the corresponding author upon request. Some generated video samples that represent the findings of this research are provided at:
\noindent https://github.com/Tasmiah1408028/Motion-Transfer-Videos

\section*{Appendix: Model Architectures}

\subsection*{A. Motion Transfer Backbone (FOMM)}

For motion transfer, we utilize the original implementation of FOMM and retain its default configurations for our experiments. For architectural and implementation details, please refer to the supplementary material of \cite{siarohin2019first}.

\subsection*{B. Keypoint Predictor Architectures}

VRNN and GRU-NF are our primary models to forecast future keypoint sequences in both single sample and diverse sample video prediction settings. The predicted sequences are used to drive FOMM for future video generation. The architectures of these two models are described as below.

\subsubsection*{B.1 VRNN}

The VRNN model is based on a single-layer GRU with 64-256 hidden units that captures temporal dependencies in 60-dimensional keypoint sequences. It integrates latent-variable modeling using a VAE-like structure. At each timestep, a 15 dimensional latent vector is sampled from a Gaussian distribution using prior and posterior networks, both implemented as feedforward layers with 512 hidden units. The latent code is concatenated with the GRU hidden state and passed through a decoder with a 512-unit hidden layer followed by a linear layer mapping to 60-dimensional keypoints. 


\subsubsection*{B.2 GRU-NF}

We use a probabilistic sequential generative model that combines a three-layer GRU network with 512-2048 hidden units per layer, followed by an NF component implemented using 12 RealNVP blocks. The GRU encodes the 60-dimensional input into a hidden representation which is then used to generate a sequence of 60-dimensional latent variables through the RealNVP blocks. Each block contains a pair of scale and translation subnetworks, each composed of five fully connected layers with 2048 hidden units and Tanh/ReLU activations, respectively. These blocks are interleaved with Batch Normalizing layers to stabilize training. 

\bibliographystyle{unsrt}
\bibliography{references}

@ARTICLE{Oprea2020,
  author  = {Oprea, S. and Martinez-Gonzalez, P. and Garcia-Garcia, A. and Castro-Vargas, J.A. and Orts-Escolano, S. and Garcia-Rodriguez, J. and Argyros, A.},
  title   = {A Review on Deep Learning Techniques for Video Prediction},
  journal = {IEEE Transactions on Pattern Analysis and Machine Intelligence},
  volume  = {44},
  number  = {6},
  pages   = {2806--2826},
  year    = {2020}
}

@ARTICLE{Shrivastava2021,
  author  = {Shrivastava, G. and Shrivastava, A.},
  title   = {Diverse Video Generation Using a Gaussian Process Trigger},
  journal = {arXiv preprint},
  eprint  = {arXiv:2107.04619},
  year    = {2021}
}

@ARTICLE{Luc2020,
  author  = {Luc, P. and Clark, A. and Dieleman, S. and Casas, D.D. and Doron, Y. and Cassirer, A. and Simonyan, K.},
  title   = {Transformation-Based Adversarial Video Prediction on Large-Scale Data},
  journal = {arXiv preprint},
  eprint  = {arXiv:2003.04035},
  year    = {2020},
  month   = {March}
}

@ARTICLE{Mathieu2015,
  author  = {Mathieu, M. and Couprie, C. and LeCun, Y.},
  title   = {Deep Multi-Scale Video Prediction Beyond Mean Square Error},
  journal = {arXiv preprint},
  eprint  = {arXiv:1511.05440},
  year    = {2015},
  month   = {November}
}

@INPROCEEDINGS{Reda2018,
  author    = {Reda, F.A. and Liu, G. and Shih, K.J. and Kirby, R. and Barker, J. and Tarjan, D. and Tao, A. and Catanzaro, B.},
  title     = {SDC-Net: Video Prediction Using Spatially-Displaced Convolution},
  booktitle = {Proceedings of the European Conference on Computer Vision (ECCV)},
  pages     = {718--733},
  year      = {2018}
}

@INPROCEEDINGS{Lopez1995,
  author    = {Lopez, R. and Huang, T.S.},
  title     = {Head Pose Computation for Very Low Bit-Rate Video Coding},
  booktitle = {International Conference on Computer Analysis of Images and Patterns},
  pages     = {440--447},
  year      = {1995},
  month     = {September},
  publisher = {Springer Berlin Heidelberg},
  address   = {Berlin, Heidelberg}
}

@ARTICLE{Koufakis1999,
  author  = {Koufakis, I. and Buxton, B.F.},
  title   = {Very Low Bit Rate Face Video Compression Using Linear Combination of 2D Face Views and Principal Components Analysis},
  journal = {Image and Vision Computing},
  volume  = {17},
  number  = {14},
  pages   = {1031--1051},
  year    = {1999}
}

@INPROCEEDINGS{Tang2019,
  author    = {Tang, J. and Hu, H. and Zhou, Q. and Shan, H. and Tian, C. and Quek, T.Q.},
  title     = {Pose Guided Global and Local GAN for Appearance Preserving Human Video Prediction},
  booktitle = {2019 IEEE International Conference on Image Processing (ICIP)},
  pages     = {614--618},
  year      = {2019},
  month     = {September},
  publisher = {IEEE}
}

@inproceedings{Villegas2018,
  title={Hierarchical long-term video prediction without supervision},
  author={Villegas, Ruben and Erhan, Dumitru and Lee, Honglak and others},
  booktitle={International Conference on Machine Learning},
  pages={6038--6046},
  year={2018},
  organization={PMLR}
}

@inproceedings{Villegas2017,
  title={Learning to generate long-term future via hierarchical prediction},
  author={Villegas, Ruben and Yang, Jimei and Zou, Yuliang and Sohn, Sungryull and Lin, Xunyu and Lee, Honglak},
  booktitle={international conference on machine learning},
  pages={3560--3569},
  year={2017},
  organization={PMLR}
}

@ARTICLE{Ranzato2014,
  author  = {Ranzato, M. and Szlam, A. and Bruna, J. and Mathieu, M. and Collobert, R. and Chopra, S.},
  title   = {Video (Language) Modeling: A Baseline for Generative Models of Natural Videos},
  journal = {arXiv preprint},
  eprint  = {arXiv:1412.6604},
  year    = {2014},
  month   = {December}
}

@inproceedings{Terwilliger2019,
  title={Recurrent flow-guided semantic forecasting},
  author={Terwilliger, Adam and Brazil, Garrick and Liu, Xiaoming},
  booktitle={2019 IEEE Winter Conference on Applications of Computer Vision (WACV)},
  pages={1703--1712},
  year={2019},
  organization={IEEE}
}

@ARTICLE{Hochreiter1997,
  author  = {Hochreiter, S. and Schmidhuber, J.},
  title   = {Long Short-Term Memory},
  journal = {Neural Computation},
  volume  = {9},
  number  = {8},
  pages   = {1735--1780},
  year    = {1997}
}

@INPROCEEDINGS{Chan2019,
  author    = {Chan, C. and Ginosar, S. and Zhou, T. and Efros, A.A.},
  title     = {Everybody Dance Now},
  booktitle = {Proceedings of the IEEE/CVF International Conference on Computer Vision (ICCV)},
  pages     = {5933--5942},
  year      = {2019}
}

@INPROCEEDINGS{Bai2024,
  author    = {Bai, X. and Haque, T. and Mohan, S. and Cai, Y. and Jeong, B. and Halasz, A. and Das, S.},
  title     = {Enhancing Bandwidth Efficiency for Video Motion Transfer Applications Using Deep Learning Based Keypoint Prediction},
  booktitle = {International Conference on Engineering Applications of Neural Networks},
  pages     = {134--151},
  year      = {2024},
  month     = {June},
  publisher = {Springer Nature Switzerland},
  address   = {Cham}
}

@ARTICLE{Aigner2018,
  author  = {Aigner, S. and Körner, M.},
  title   = {FutureGAN: Anticipating the future frames of video sequences using spatio-temporal 3D convolutions in progressively growing GANs},
  journal={arXiv preprint},
  eprint={arXiv:1810.01325},
  year = {2018}
}

@INPROCEEDINGS{Shi2015,
  author    = {Shi, Xingjian and Chen, Zhourong and Wang, Hao and Yeung, Dit-Yan and Wong, Wai-Kin and Woo, Wang-Chun},
  title     = {Convolutional LSTM Network: A Machine Learning Approach for Precipitation Nowcasting},
  booktitle = {Advances in Neural Information Processing Systems (NeurIPS)},
  volume    = {28},
  year      = {2015}
}

@INPROCEEDINGS{Wang2018,
  author    = {Wang, Y. and Jiang, L. and Yang, M.H. and Li, L.J. and Long, M. and Fei-Fei, L.},
  title     = {Eidetic 3D LSTM: A Model for Video Prediction and Beyond},
  booktitle = {International Conference on Learning Representations (ICLR)},
  year      = {2018},
  month     = {September}
}

@ARTICLE{Villar-Corrales2022,
  author  = {Villar-Corrales, Angel and Karapetyan, Ani and Boltres, Andreas and Behnke, Sven},
  title   = {MSPred: Video Prediction at Multiple Spatio-Temporal Scales with Hierarchical Recurrent Networks},
  journal = {arXiv preprint},
  eprint  = {arXiv:2203.09303},
  year    = {2022}
}

@INPROCEEDINGS{Shi2022,
  author    = {Shi, Zhihao and Xu, Xiangyu and Liu, Xiaohong and Chen, Jun and Yang, Ming-Hsuan},
  title     = {Video Frame Interpolation Transformer},
  booktitle = {Proceedings of the IEEE/CVF Conference on Computer Vision and Pattern Recognition (CVPR)},
  pages     = {17482--17491},
  year      = {2022}
}

@ARTICLE{Ye2023,
  author  = {Ye, Xi and Bilodeau, Guillaume-Alexandre},
  title   = {Video Prediction by Efficient Transformers},
  journal = {Image and Vision Computing},
  volume  = {130},
  pages   = {104612},
  year    = {2023}
}

@INPROCEEDINGS{Lu2022,
  author    = {Lu, L. and Wu, R. and Lin, H. and Lu, J. and Jia, J.},
  title     = {Video Frame Interpolation with Transformer},
  booktitle = {Proceedings of the IEEE/CVF Conference on Computer Vision and Pattern Recognition (CVPR)},
  pages     = {3532--3542},
  year      = {2022}
}

@INPROCEEDINGS{Minderer2019,
  author    = {Minderer, M. and Sun, C. and Villegas, R. and Cole, F. and Murphy, K.P. and Lee, H.},
  title     = {Unsupervised Learning of Object Structure and Dynamics from Videos},
  booktitle = {Advances in Neural Information Processing Systems (NeurIPS)},
  volume    = {32},
  year      = {2019}
}

@ARTICLE{Walker2021,
  author  = {Walker, J. and Razavi, A. and van den Oord, A.},
  title   = {Predicting Video with VQVAE},
  journal = {arXiv preprint},
  eprint  = {arXiv:2103.01950},
  year    = {2021}
}

@INPROCEEDINGS{Walker2017,
  author    = {Walker, J. and Marino, K. and Gupta, A. and Hebert, M.},
  title     = {The Pose Knows: Video Forecasting by Generating Pose Futures},
  booktitle = {Proceedings of the IEEE International Conference on Computer Vision (ICCV)},
  pages     = {3332--3341},
  year      = {2017}
}

@INPROCEEDINGS{Gao2022,
  author    = {Gao, Z. and Tan, C. and Wu, L. and Li, S.Z.},
  title     = {SimVP: Simpler Yet Better Video Prediction},
  booktitle = {Proceedings of the IEEE/CVF Conference on Computer Vision and Pattern Recognition (CVPR)},
  pages     = {3170--3180},
  year      = {2022}
}

@INPROCEEDINGS{Xu2018,
  author    = {Xu, Z. and Wang, Y. and Long, M. and Wang, J. and Kliss, M.},
  title     = {PredCNN: Predictive Learning with Cascade Convolutions},
  booktitle = {Proceedings of the International Joint Conference on Artificial Intelligence (IJCAI)},
  pages     = {2940--2947},
  year      = {2018},
  month     = {July}
}

@ARTICLE{Zhou2021,
  author  = {Zhou, Y. and Luo, C. and Sun, X. and Zha, Z.J. and Zeng, W.},
  title   = {VAE\textsuperscript{2}: Preventing Posterior Collapse of Variational Video Predictions in the Wild},
  journal = {arXiv preprint},
  eprint  = {arXiv:2101.12050},
  year    = {2021}
}

@ARTICLE{Ming2024,
  author  = {Ming, R. and Huang, Z. and Ju, Z. and Hu, J. and Peng, L. and Zhou, S.},
  title   = {A Survey on Video Prediction: From Deterministic to Generative Approaches},
  journal = {arXiv preprint},
  eprint  = {arXiv:2401.14718},
  year    = {2024}
}

@INPROCEEDINGS{Jang2018,
  author    = {Jang, Yunseok and Kim, Gunhee and Song, Yale},
  title     = {Video Prediction with Appearance and Motion Conditions},
  booktitle = {Proceedings of the International Conference on Machine Learning (ICML)},
  year      = {2018}
}

@INPROCEEDINGS{Esser2021,
  author    = {Esser, Patrick and Rombach, Robin and Ommer, Bjorn},
  title     = {Taming Transformers for High-Resolution Image Synthesis},
  booktitle = {Proceedings of the IEEE/CVF Conference on Computer Vision and Pattern Recognition (CVPR)},
  year      = {2021}
}

@inproceedings{Rezende2015,
  title={Variational inference with normalizing flows},
  author={Rezende, Danilo and Mohamed, Shakir},
  booktitle={International conference on machine learning},
  pages={1530--1538},
  year={2015},
  organization={PMLR}
}

@ARTICLE{Papamakarios2021,
  author  = {Papamakarios, G. and Nalisnick, E. and Rezende, D.J. and Mohamed, S. and Lakshminarayanan, B.},
  title   = {Normalizing Flows for Probabilistic Modeling and Inference},
  journal = {Journal of Machine Learning Research},
  volume  = {22},
  number  = {57},
  pages   = {1--64},
  year    = {2021}
}

@ARTICLE{Dinh2016,
  author  = {Dinh, L. and Sohl-Dickstein, J. and Bengio, S.},
  title   = {Density Estimation Using Real NVP},
  journal = {arXiv preprint},
  eprint  = {arXiv:1605.08803},
  year    = {2016}
}

@ARTICLE{Rasul2020,
  author  = {Rasul, K. and Sheikh, A.S. and Schuster, I. and Bergmann, U. and Vollgraf, R.},
  title   = {Multivariate Probabilistic Time Series Forecasting via Conditioned Normalizing Flows},
  journal = {arXiv preprint},
  eprint  = {arXiv:2002.06103},
  year    = {2020}
}

@INPROCEEDINGS{Xue2016,
  author    = {Xue, T. and Wu, J. and Bouman, K. and Freeman, B.},
  title     = {Visual Dynamics: Probabilistic Future Frame Synthesis via Cross Convolutional Networks},
  booktitle = {Advances in Neural Information Processing Systems (NeurIPS)},
  volume    = {29},
  year      = {2016}
}

@INPROCEEDINGS{Goroshin2015,
  author    = {Goroshin, R. and Mathieu, M.F. and LeCun, Y.},
  title     = {Learning to Linearize Under Uncertainty},
  booktitle = {Advances in Neural Information Processing Systems (NeurIPS)},
  volume    = {28},
  year      = {2015}
}

@ARTICLE{Fragkiadaki2017,
  author  = {Fragkiadaki, K. and Huang, J. and Alemi, A. and Vijayanarasimhan, S. and Ricco, S. and Sukthankar, R.},
  title   = {Motion Prediction Under Multimodality with Conditional Stochastic Networks},
  journal = {arXiv preprint},
  eprint = {arXiv:1705.02082},
  year    = {2017}
}

@ARTICLE{Babaeizadeh2017,
  author  = {Babaeizadeh, M. and Finn, C. and Erhan, D. and Campbell, R.H. and Levine, S.},
  title   = {Stochastic Variational Video Prediction},
  journal = {arXiv preprint},
  eprint  = {arXiv:1710.11252},
  year    = {2017}
}

@INPROCEEDINGS{Denton2018,
  title={Stochastic video generation with a learned prior},
  author={Denton, Emily and Fergus, Rob},
  booktitle={International conference on machine learning},
  pages={1174--1183},
  year={2018},
  organization={PMLR}
}

@ARTICLE{Lee2018,
  author  = {Lee, A.X. and Zhang, R. and Ebert, F. and Abbeel, P. and Finn, C. and Levine, S.},
  title   = {Stochastic Adversarial Video Prediction},
  journal = {arXiv preprint},
  eprint  = {arXiv:1804.01523},
  year    = {2018}
}

@INPROCEEDINGS{Sonderby2016,
  author    = {Sønderby, C.K. and Raiko, T. and Maaløe, L. and Sønderby, S.K. and Winther, O.},
  title     = {Ladder Variational Autoencoders},
  booktitle = {Advances in Neural Information Processing Systems (NeurIPS)},
  volume    = {29},
  year      = {2016}
}

@INPROCEEDINGS{Gur2020,
  author    = {Gur, S. and Benaim, S. and Wolf, L.},
  title     = {Hierarchical Patch VAE-GAN: Generating Diverse Videos from a Single Sample},
  booktitle = {Advances in Neural Information Processing Systems (NeurIPS)},
  volume    = {33},
  pages     = {16761--16772},
  year      = {2020}
}

@ARTICLE{Yang2019,
  author  = {Yang, D. and Hong, S. and Jang, Y. and Zhao, T. and Lee, H.},
  title   = {Diversity-Sensitive Conditional Generative Adversarial Networks},
  journal = {arXiv preprint},
  eprint  = {arXiv:1901.09024},
  year    = {2019}
}

@ARTICLE{Pottorff2019,
  author  = {Pottorff, R. and Nielsen, J. and Wingate, D.},
  title   = {Video Extrapolation with an Invertible Linear Embedding},
  journal = {arXiv preprint},
  eprint= {arXiv:1903.00133},
  year = {2019}
}

@ARTICLE{Kumar2019,
  author  = {Kumar, M. and Babaeizadeh, M. and Erhan, D. and Finn, C. and Levine, S. and Dinh, L. and Kingma, D.},
  title   = {VideoFlow: A Conditional Flow-Based Model for Stochastic Video Generation},
  journal = {arXiv preprint},
  eprint  = {arXiv:1903.01434},
  year    = {2019}
}

@ARTICLE{Ma2020,
  author  = {Ma, Y.J. and Inala, J.P. and Jayaraman, D. and Bastani, O.},
  title   = {Diverse Sampling for Normalizing Flow Based Trajectory Forecasting},
  journal = {arXiv preprint},
  eprint  = {arXiv:2011.15084},
  year    = {2020},
  number  = {7(8)}
}

@inproceedings{Hu2020,
  title={Probabilistic future prediction for video scene understanding},
  author={Hu, Anthony and Cotter, Fergal and Mohan, Nikhil and Gurau, Corina and Kendall, Alex},
  booktitle={European Conference on Computer Vision},
  pages={767--785},
  year={2020},
  organization={Springer}
}

@INPROCEEDINGS{Wang2021,
  author    = {Wang, T.C. and Mallya, A. and Liu, M.Y.},
  title     = {One-Shot Free-View Neural Talking-Head Synthesis for Video Conferencing},
  booktitle = {Proceedings of the IEEE/CVF Conference on Computer Vision and Pattern Recognition (CVPR)},
  pages     = {10039--10049},
  year      = {2021}
}

@ARTICLE{Yang2022,
  author  = {Yang, H.C. and Rahmanti, A.R. and Huang, C.W. and Li, Y.C.},
  title   = {How Can Research on Artificial Empathy Be Enhanced by Applying Deepfakes?},
  journal = {Journal of Medical Internet Research},
  volume  = {24},
  number  = {3},
  pages   = {e29506},
  year    = {2022},
  month   = {March}
}

@inproceedings{Rehman2015,
  title={Display device-adapted video quality-of-experience assessment},
  author={Rehman, Abdul and Zeng, Kai and Wang, Zhou},
  booktitle={Human vision and electronic imaging XX},
  volume={9394},
  pages={27--37},
  year={2015},
  organization={SPIE}
}

@INPROCEEDINGS{Kotevski2009,
  author    = {Kotevski, Z. and Mitrevski, P.},
  title     = {Experimental Comparison of PSNR and SSIM Metrics for Video Quality Estimation},
  booktitle = {International Conference on ICT Innovations},
  pages     = {357--366},
  year      = {2009},
  month     = {September},
  publisher = {Springer Berlin Heidelberg},
  address   = {Berlin, Heidelberg}
}

@article{Wang2004SSIM,
  author={Wang, Zhou and Bovik, Alan C. and Sheikh, Hamid R. and Simoncelli, Eero P.},
  journal={IEEE Transactions on Image Processing}, 
  title={Image Quality Assessment: From Error Visibility to Structural Similarity}, 
  year={2004},
  volume={13},
  number={4},
  pages={600-612},
  doi={10.1109/TIP.2003.819861}
}

@article{unterthiner2018fvd,
  title={Towards Accurate Generative Models of Video: A New Metric \& Challenges},
  author={Unterthiner, Thomas and van den Oord, A{\"a}ron and Heusel, Mario and Ramsauer, Hubert and Nessler, Bernhard and Hochreiter, Sepp},
  journal={NeurIPS Workshop on Challenges and Opportunities for Deep Learning in Autonomous Driving},
  year={2018},
  eprint={arXiv:1812.01717},
  primaryClass={cs.LG}
}

@inproceedings{Ullah2020,
  title={Exploring clinical time series forecasting with meta-features in variational recurrent models},
  author={Ullah, Sibghat and Xu, Zhao and Wang, Hao and Menzel, Stefan and Sendhoff, Bernhard and B{\"a}ck, Thomas},
  booktitle={2020 international joint conference on neural networks (IJCNN)},
  pages={1--9},
  year={2020},
  organization={IEEE}
}

@inproceedings{rezende2014stochastic,
  title={Stochastic Backpropagation and Approximate Inference in Deep Generative Models},
  author={Rezende, Danilo Jimenez and Mohamed, Shakir and Wierstra, Daan},
  booktitle={Proceedings of the 31st International Conference on Machine Learning (ICML)},
  year={2014},
  journal={arXiv preprint},
  eprint={arXiv:1401.4082}
}

@INPROCEEDINGS{Zakharov2019,
  author    = {Zakharov, E. and Shysheya, A. and Burkov, E. and Lempitsky, V.},
  title     = {Few-Shot Adversarial Learning of Realistic Neural Talking Head Models},
  booktitle = {Proceedings of the IEEE/CVF International Conference on Computer Vision (ICCV)},
  pages     = {9459--9468},
  year      = {2019}
}

@article{cho2014gru,
  title={Empirical Evaluation of Gated Recurrent Neural Networks on Sequence Modeling},
  author={Cho, Kyunghyun and Van Merriënboer, Bart and Gulcehre, Caglar and Bahdanau, Dzmitry and Bougares, Fethi and Schwenk, Holger and Bengio, Yoshua},
  journal={arXiv preprint},
  eprint = {arXiv:1406.1078},
  year={2014}
}

@INPROCEEDINGS{Chung2015,
  author    = {Chung, J. and Kastner, K. and Dinh, L. and Goel, K. and Courville, A.C. and Bengio, Y.},
  title     = {A Recurrent Latent Variable Model for Sequential Data},
  booktitle = {Advances in Neural Information Processing Systems (NeurIPS)},
  volume    = {28},
  year      = {2015}
}

@article{siarohin2019first,
  title={First order motion model for image animation},
  author={Siarohin, Aliaksandr and Lathuili{\`e}re, St{\'e}phane and Tulyakov, Sergey and Ricci, Elisa and Sebe, Nicu},
  journal={Advances in neural information processing systems},
  volume={32},
  year={2019},
  url={https://proceedings.neurips.cc/paper_files/paper/2019/hash/31c0b36aef265d9221af80872ceb62f9-Abstract.html}
}

@INPROCEEDINGS{Zhao2022,
  author    = {Zhao, J. and Zhang, H.},
  title     = {Thin-Plate Spline Motion Model for Image Animation},
  booktitle = {Proceedings of the IEEE/CVF Conference on Computer Vision and Pattern Recognition (CVPR)},
  pages     = {3657--3666},
  year      = {2022}
}

@ARTICLE{Yesiltepe2024,
  author  = {Yesiltepe, H. and Meral, T.H.S. and Dunlop, C. and Yanardag, P.},
  title   = {MotionShop: Zero-Shot Motion Transfer in Video Diffusion Models with Mixture of Score Guidance},
  journal = {arXiv preprint},
  eprint  = {arXiv:2412.05355},
  year    = {2024}
}

@ARTICLE{Meral2024,
  author  = {Meral, T.H.S. and Yesiltepe, H. and Dunlop, C. and Yanardag, P.},
  title   = {MotionFlow: Attention-Driven Motion Transfer in Video Diffusion Models},
  journal = {arXiv preprint},
  eprint  = {arXiv:2412.05275},
  year    = {2024}
}

@ARTICLE{Pondaven2024,
  author  = {Pondaven, A. and Siarohin, A. and Tulyakov, S. and Torr, P. and Pizzati, F.},
  title   = {Video Motion Transfer with Diffusion Transformers},
  journal = {arXiv preprint},
  eprint  = {arXiv:2412.07776},
  year    = {2024}
}

@article{kingma2013auto,
  title={Auto-encoding variational bayes},
  author={Kingma, Diederik P},
  journal={arXiv preprint},
  eprint = {arXiv:1312.6114},
  year={2013}
}

@article{chen2024mixed,
  title={Mixed Gaussian Flow for Diverse Trajectory Prediction},
  author={Chen, Jiahe and Cao, Jinkun and Lin, Dahua and Kitani, Kris and Pang, Jiangmiao},
  journal={arXiv preprint}, eprint = {arXiv:2402.12238},
  year={2024}
}

@inproceedings{siarohin2019animating,
  title={Animating arbitrary objects via deep motion transfer},
  author={Siarohin, Aliaksandr and Lathuili{\`e}re, St{\'e}phane and Tulyakov, Sergey and Ricci, Elisa and Sebe, Nicu},
  booktitle={Proceedings of the IEEE/CVF Conference on Computer Vision and Pattern Recognition},
  pages={2377--2386},
  year={2019}
}

@article{ebert2017self,
  title={Self-Supervised Visual Planning with Temporal Skip Connections.},
  author={Ebert, Frederik and Finn, Chelsea and Lee, Alex X and Levine, Sergey},
  journal={CoRL},
  volume={12},
  number={16},
  pages={23},
  year={2017}
}

@article{nagrani2017voxceleb,
  title={Voxceleb: a large-scale speaker identification dataset},
  author={Nagrani, Arsha and Chung, Joon Son and Zisserman, Andrew},
  journal={arXiv preprint}, eprint = {arXiv:1706.08612},
  year={2017}
}

@article{luo2024beyond,
  title={Beyond FVD: Enhanced Evaluation Metrics for Video Generation Quality},
  author={Luo, Ge Ya and Favero, Gian Mario and Luo, Zhi Hao and Jolicoeur-Martineau, Alexia and Pal, Christopher},
  journal={arXiv preprint},
  eprint = {arXiv:2410.05203},
  year={2024}
}

@article{gretton2012kernel,
  title={A kernel two-sample test},
  author={Gretton, Arthur and Borgwardt, Karsten M and Rasch, Malte J and Sch{\"o}lkopf, Bernhard and Smola, Alexander},
  journal={The Journal of Machine Learning Research},
  volume={13},
  number={1},
  pages={723--773},
  year={2012},
  publisher={JMLR. org}
}

@inproceedings{lu2023transflow,
  title={Transflow: Transformer as flow learner},
  author={Lu, Yawen and Wang, Qifan and Ma, Siqi and Geng, Tong and Chen, Yingjie Victor and Chen, Huaijin and Liu, Dongfang},
  booktitle={Proceedings of the IEEE/CVF conference on computer vision and pattern recognition},
  pages={18063--18073},
  year={2023}
}

@inproceedings{lu2024promotion,
  title={Promotion: Prototypes as motion learners},
  author={Lu, Yawen and Liu, Dongfang and Wang, Qifan and Han, Cheng and Cui, Yiming and Cao, Zhiwen and Zhang, Xueling and Chen, Yingjie Victor and Fan, Heng},
  booktitle={Proceedings of the IEEE/CVF Conference on Computer Vision and Pattern Recognition},
  pages={28109--28119},
  year={2024}
}

@article{han2024prototypical,
  title={Prototypical transformer as unified motion learners},
  author={Han, Cheng and Lu, Yawen and Sun, Guohao and Liang, James C and Cao, Zhiwen and Wang, Qifan and Guan, Qiang and Dianat, Sohail A and Rao, Raghuveer M and Geng, Tong and others},
  journal={arXiv preprint arXiv:2406.01559},
  year={2024}
}

@article{liu2020video,
  title={Video object detection for autonomous driving: Motion-aid feature calibration},
  author={Liu, Dongfang and Cui, Yiming and Chen, Yingjie and Zhang, Jiyong and Fan, Bin},
  journal={Neurocomputing},
  volume={409},
  pages={1--11},
  year={2020},
  publisher={Elsevier}
}

@inproceedings{thewlis2017unsupervised,
  title={Unsupervised learning of object landmarks by factorized spatial embeddings},
  author={Thewlis, James and Bilen, Hakan and Vedaldi, Andrea},
  booktitle={Proceedings of the IEEE international conference on computer vision},
  pages={5916--5925},
  year={2017}
}

@inproceedings{zhang2018unsupervised,
  title={Unsupervised discovery of object landmarks as structural representations},
  author={Zhang, Yuting and Guo, Yijie and Jin, Yixin and Luo, Yijun and He, Zhiyuan and Lee, Honglak},
  booktitle={Proceedings of the IEEE Conference on Computer Vision and Pattern Recognition},
  pages={2694--2703},
  year={2018}
}

@article{guo2024liveportrait,
  title={Liveportrait: Efficient portrait animation with stitching and retargeting control},
  author={Guo, Jianzhu and Zhang, Dingyun and Liu, Xiaoqiang and Zhong, Zhizhou and Zhang, Yuan and Wan, Pengfei and Zhang, Di},
  journal={arXiv preprint arXiv:2407.03168},
  year={2024}
}

@inproceedings{wang2024continuous,
  title={Continuous piecewise-affine based motion model for image animation},
  author={Wang, Hexiang and Liu, Fengqi and Zhou, Qianyu and Yi, Ran and Tan, Xin and Ma, Lizhuang},
  booktitle={Proceedings of the AAAI Conference on Artificial Intelligence},
  volume={38},
  number={6},
  pages={5427--5435},
  year={2024}
}

@article{bommasani2021opportunities,
  title={On the opportunities and risks of foundation models},
  author={Bommasani, Rishi et al.},
  journal={arXiv preprint arXiv:2108.07258},
  year={2021}
}

@inproceedings{kirillov2023segment,
  title={Segment anything},
  author={Kirillov, Alexander and Mintun, Eric and Ravi, Nikhila and Mao, Hanzi and Rolland, Chloe and Gustafson, Laura and Xiao, Tete and Whitehead, Spencer and Berg, Alexander C and Lo, Wan-Yen and others},
  booktitle={Proceedings of the IEEE/CVF international conference on computer vision},
  pages={4015--4026},
  year={2023}
}

@inproceedings{tumanyan2022splicing,
  title={Splicing vit features for semantic appearance transfer},
  author={Tumanyan, Narek and Bar-Tal, Omer and Bagon, Shai and Dekel, Tali},
  booktitle={Proceedings of the IEEE/CVF Conference on Computer Vision and Pattern Recognition},
  pages={10748--10757},
  year={2022}
}

@software{fvcore,
  author       = {{Facebook AI Research (FAIR)}},
  title        = {fvcore: A light-weight core library for computer vision},
  year         = {2022},
  howpublished = {\url{https://github.com/facebookresearch/fvcore}},
  note         = {Apache-2.0 license. Common utilities used by Detectron2, PySlowFast, etc.},
}

@misc{zoom_system_requirements_desktop,
  author       = {{Zoom Video Communications, Inc.}},
  title        = {Zoom system requirements: Windows, macOS, Linux},
  howpublished = {\url{https://support.zoom.com/hc/en/article?id=zm_kb&sysparm_article=KB0060748}},
  note         = {Zoom Support article. Accessed 2025-10-13},
  year         = {n.d.},
  urldate      = {2025-10-13}
}

@misc{webex_min_bandwidth_meetings,
  author       = {{Cisco Webex}},
  title        = {What are the Minimum Bandwidth Requirements for Sending and Receiving Video in Cisco Webex Meetings?},
  howpublished = {\url{https://help.webex.com/en-us/article/WBX22158/What-are-the-Minimum-Bandwidth-Requirements-for-Sending-and-Receiving-Video-inCisco-Webex-Meetings}},
  note         = {Webex Help Center article. Last updated Dec 25, 2023. Accessed 2025-10-13},
  year         = {2023},
  urldate      = {2025-10-13}
}

@inproceedings{gao2024empirical,
  title={An empirical study on low gpu utilization of deep learning jobs},
  author={Gao, Yanjie and He, Yichen and Li, Xinze and Zhao, Bo and Lin, Haoxiang and Liang, Yoyo and Zhong, Jing and Zhang, Hongyu and Wang, Jingzhou and Zeng, Yonghua and others},
  booktitle={Proceedings of the IEEE/ACM 46th International Conference on Software Engineering},
  pages={1--13},
  year={2024}
}

\end{document}